\documentclass{article}

\usepackage{amsmath, amssymb, amsthm}
\usepackage[final,main]{neurips_2026}

\usepackage[utf8]{inputenc}
\usepackage[T1]{fontenc}
\DeclareUnicodeCharacter{00B1}{\ensuremath{\pm}}
\DeclareUnicodeCharacter{00D7}{\ensuremath{\times}}
\DeclareUnicodeCharacter{2013}{--}
\DeclareUnicodeCharacter{2014}{---}
\DeclareUnicodeCharacter{2192}{\ensuremath{\rightarrow}}
\usepackage[hypertexnames=false]{hyperref}
\usepackage{url}
\usepackage{booktabs}
\usepackage{multirow}
\usepackage{amsfonts}
\usepackage{nicefrac}
\usepackage[activate={true,nocompatibility},final,tracking=true,patch=none]{microtype}
\usepackage{xcolor}
\usepackage{tikz}
\usepackage{pgfplots}

\makeatletter
\pgfmathdeclarefunction*{thisrow}{1}{}
\pgfmathdeclarefunction*{thisrowno}{1}{}
\def\pgfplots@define@dummies@for@pseudoconstants{}
\def\pgfmathredeclarefunction#1#2{%
  \expandafter\ifx\csname pgfmath@function@#1\endcsname\relax
    \pgfmath@error{The function `#1' does not exist}{}%
  \else
    \expandafter\def\csname pgfmath#1@\endcsname##1{#2}%
  \fi}
\makeatother
\usepackage{graphicx}
\usepackage{wrapfig}
\graphicspath{{figures/}}
\pgfplotsset{compat=1.18}
\usetikzlibrary{arrows,positioning,shapes,calc,patterns,decorations.pathreplacing}
\usepackage{enumitem}
\usepackage{placeins}
\usepackage{mathtools}
\usepackage{algorithm}
\usepackage{algpseudocode}
\usepackage{color}
\usepackage{natbib}
\usepackage{comment}
\newcommand{\cmark}{\checkmark}

\newtheorem{theorem}{Theorem}[section]

\newtheorem{proposition}[theorem]{Proposition}

\newtheorem{definition}[theorem]{Definition}

\newtheorem{remark}[theorem]{Remark}

\DeclareMathOperator{\Var}{Var}

\DeclareMathOperator{\KL}{KL}

\newcommand{\E}{\mathbb{E}}

\newcommand{\Prob}{\mathbb{P}}

\newcommand{\cY}{\mathcal{Y}}

\newcommand{\cH}{\mathcal{H}}

\newcommand{\cL}{\mathcal{L}}

\title{Margin Dynamics for Large Language Model Alignment}

\author{%
  Xingzi Xu\\
  Amazon\\
  \texttt{xingzixu@amazon.com}\\
  \And
  Mehmet Saygin Seyfioglu\\
  Amazon\\
  \texttt{mseyfiog@amazon.com}\\
  \And
  Karim Bouyarmane \\
  Amazon \\
  \texttt{bouykari@amazon.com} \\
}

\begin{document}

\maketitle

\begin{abstract}
Practitioners choose among LLM alignment methods largely by trial and error.
We provide a theoretical foundation: each pairwise method defines a vector field on the probability simplex that governs how the model redistributes mass between preferred and dispreferred completions. Projecting onto a single preferred/dispreferred pair reduces this to a one-dimensional margin ODE with a closed-form solution.
These solutions reveal that DPO margins grow without bound (logarithmically as $O(\log t)$ in the gradient-flow limit) while IPO margins converge to a finite $1/(2\tau)$; that label noise induces a qualitative phase transition in DPO (unbounded $\to$ bounded) but affects IPO only quantitatively; and that gradient concentration under asymmetric weighting governs trainability at scale.
We confirm all three predictions across five model families spanning four benchmarks.
The closed-form solutions also enable principled method design: we prove that adding \emph{any} label smoothing $\varepsilon > 0$ to SimPO creates a finite margin attractor at $\Delta^* = \beta^{-1}\log((1{-}\varepsilon)/\varepsilon)$, while $\varepsilon{=}0$ (standard SimPO) has no attractor.
We test this across three models (8B--9B) and confirm the prediction: the theory-derived configuration (TPO-S, $\varepsilon{=}0.05$) is competitive with the strongest baselines, attaining the highest mean IFEval score among the compared methods on all three, validating that the predicted finite attractor is a usable operating point rather than proposing a new state-of-the-art loss.
\end{abstract}

\section{Introduction}
\label{sec:intro}

The alignment toolkit has grown rapidly: supervised fine-tuning (SFT; \citealt{ouyang2022rlhf}), reinforcement learning from human feedback (PPO; \citealt{schulman2017ppo}), and a family of offline preference methods that bypass reward modeling by optimizing pairwise losses directly (DPO, \citealt{rafailov2023dpo}; IPO, \citealt{azar2024general}; SimPO, \citealt{meng2024simpo}; KTO, \citealt{ethayarajh2024kto}; GRPO, \citealt{shao2024grpo}). Each introduces different losses, hyperparameters, and failure modes. Selecting among them remains empirical: a method that excels on one model size or task can collapse on another, and no existing theory explains why.

We close this gap by analyzing alignment as \emph{margin dynamics} on the probability simplex. Every method defines a field $F = F^+ - F^-$ governing how the model redistributes mass across completions. Projecting onto a two-completion effective theory yields margin ODEs with closed-form solutions. DPO's margin grows without bound (sub-linearly, $O(\log t)$ under gradient flow) while IPO's converges to a finite value, and a single fitted calibration parameter $\beta_{\text{eff}}$ maps the theory to full-scale training across all settings we test. Adam enters as a strictly positive time reparameterization, so we claim the functional character (unbounded vs. bounded) and the invariant set rather than the step-time rate constant, which $\beta_{\text{eff}}$ absorbs (Appendix~\ref{app:beta_eff_proof}).

The closed-form solutions organize the alignment landscape along three axes (Figure~\ref{fig:margin_dynamics}). \emph{Axis 1, Repulsion:} methods with \emph{targeted} repulsion (DPO, IPO, SimPO, KTO) explicitly push down dispreferred completions and maintain $\Omega(\pi)$ gradient pressure, while \emph{implicit-only} methods (SFT, RFT) suffer $O(\pi^2)$ self-limiting decay. \emph{Axis 2, Weighting:} the weighting function $w(\Delta)$ determines margin dynamics. Sigmoid weights (DPO, SimPO) produce unbounded margins, linear weights (IPO, PPO) converge to finite margins, and label noise induces a qualitative phase transition in the former but only a quantitative shift in the latter. \emph{Axis 3, Reference:} frozen references (DPO, IPO) cause gradient decay as the policy drifts, reference-free methods (SimPO) avoid decay but risk unbounded drift, and asymmetric weighting ($\mathrm{DPO}_\gamma$) exposes a gradient-concentration regime that governs trainability at scale. We derive quantitative predictions along each axis and confirm them across five model families (Section~\ref{sec:prospective}).

The theoretical framework also specifies desired dynamical properties and facilitates identifying losses that produce them. The noise phase transition (Theorem~\ref{thm:noise}) predicts that adding \emph{any} label smoothing $\varepsilon > 0$ to SimPO creates a stable fixed point at $\Delta^* = \beta^{-1}\log((1{-}\varepsilon)/\varepsilon)$, converting unbounded margin growth into convergent dynamics. We set $\varepsilon{=}0.05$ ($\Delta^* \approx 29$) and call this configuration TPO-S (Targeted Preference Optimization, label-Smoothed variant) for brevity; we also test a Huber-loss variant (TPO-H) derived from the gradient concentration analysis. In summary, this paper makes three contributions: (i)~a margin-dynamics framework yielding closed-form ODEs for major pairwise alignment methods, producing a three-axis taxonomy with testable predictions; (ii)~quantitative confirmation across five model families and four benchmarks; and (iii)~a constructive test in which the theory-predicted configuration (TPO-S, $\varepsilon{=}0.05$), specified a priori with no tuning, is confirmed to be a competitive operating point --- attaining the highest mean IFEval among the compared methods across three models (8B--9B) --- validating the finite-attractor prediction rather than proposing a state-of-the-art loss.

\section{Margin Dynamics and Constructive Validation}
\label{sec:theory}

Let $\pi_\theta(y|x)$ denote a language model policy parameterized by $\theta$. For a fixed prompt $x$ and completions $\cY_x = \{y_1, \ldots, y_n\}$, the policy defines a point on the probability simplex $\Delta^{n-1}$. Each training method defines a vector field $F: \Delta^{n-1} \to T\Delta^{n-1}$ governing the evolution of $\pi$, tangent to the simplex ($\sum_i F_i = 0$). We decompose $F = F^+ - F^-$ into attraction (toward desired outputs) and repulsion (away from undesired outputs), following the decomposition of \citet{deng2026drifting}. The key step is projecting this field onto a two-completion simplex $\{y_w, y_l\}$, reducing the dynamics to a one-dimensional margin ODE $\dot\Delta = w(\Delta)$ that admits closed-form solutions. Here $\Delta = \log(\pi(y_w)/\pi(y_l))$ is the log-margin between preferred and dispreferred completions, $\sigma$ denotes the sigmoid function, and $\beta$, $\tau$, $\gamma$ are method-specific parameters defined below. To ground the notation, consider DPO \citep{rafailov2023dpo}:
\begin{equation}
\label{eq:dpo}
\cL_{\text{DPO}} = -\log\sigma\bigl(\beta(\Delta_\theta - \Delta_{\text{ref}})\bigr), \quad \Delta_\theta = \log\frac{\pi_\theta(y_w|x)}{\pi_\theta(y_l|x)}, \quad \Delta_{\text{ref}} = \log\frac{\pi_{\text{ref}}(y_w|x)}{\pi_{\text{ref}}(y_l|x)},
\end{equation}
where $\beta > 0$ is the inverse temperature. More generally, weighting the rejected term by $\gamma \in [0,1]$ gives the asymmetric variant $\cL_{\text{DPO}_\gamma} = -\log\sigma(\beta\,\Delta_\gamma)$, where $\Delta_\gamma = (\log\pi_\theta(y_w|x) - \log\pi_{\text{ref}}(y_w|x)) - \gamma(\log\pi_\theta(y_l|x) - \log\pi_{\text{ref}}(y_l|x))$: $\gamma{=}0$ removes the explicit penalty on dispreferred completions, $\gamma{=}1$ recovers standard DPO. IPO \citep{azar2024general} replaces the sigmoid with a squared penalty: $\cL_{\text{IPO}} = (\Delta_\theta - \Delta_{\text{ref}} - 1/(2\tau))^2$, where $\tau$ controls regularization strength. We develop the solutions along three axes, each revealing a qualitatively different aspect of alignment dynamics.

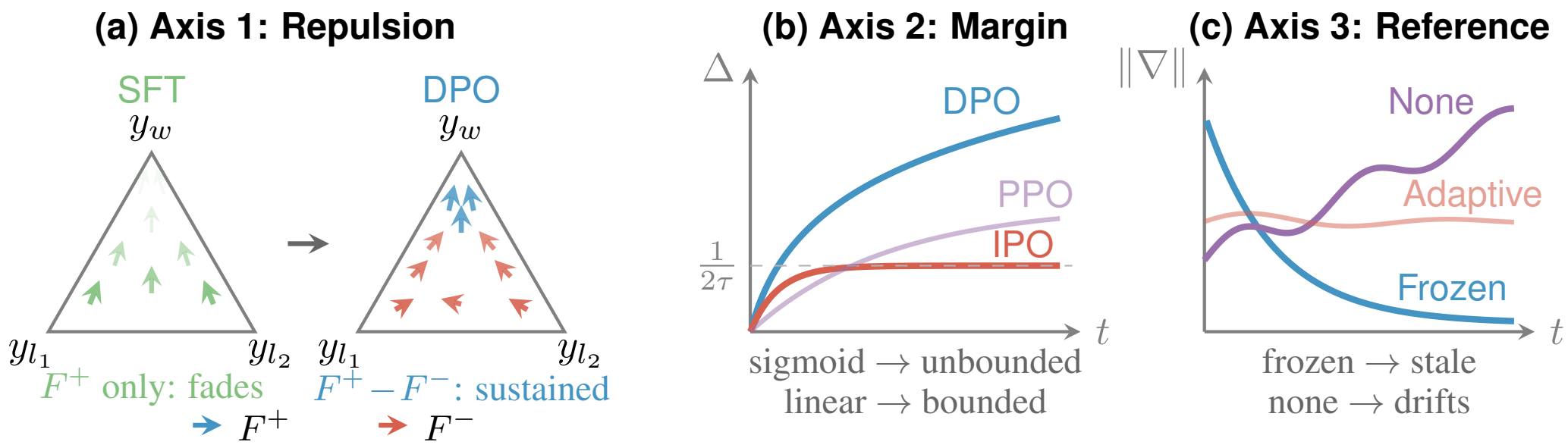
\begin{figure}[t]
\centering
\begin{tikzpicture}[>=stealth, x=0.36cm, y=0.48cm]

\definecolor{cblue}{HTML}{4393C3}
\definecolor{cred}{HTML}{D6604D}
\definecolor{cgreen}{HTML}{7FBF7B}
\definecolor{cpurp}{HTML}{9970AB}

\begin{scope}[shift={(0,0)}]
  \node[font=\sffamily\bfseries\small] at (5.5,5.5) {(a) Axis 1: Repulsion};

  \node[font=\sffamily\small, cgreen] at (2.5,4.4) {SFT};
  \draw[thick, black!50] (0,0) -- (5,0) -- (2.5,3.25) -- cycle;
  \node at (2.5,3.7) {$y_w$};
  \node at (-0.4,-0.4) {$y_{l_1}$};
  \node at (5.4,-0.4) {$y_{l_2}$};

  \foreach \x/\y/\a in {1.0/0.5/0.9, 4.0/0.5/0.9, 2.5/0.7/0.7, 1.6/1.2/0.5, 3.4/1.2/0.5, 2.5/1.8/0.2, 2.2/2.5/0.08, 2.8/2.5/0.08} {
    \pgfmathsetmacro{\dx}{2.5-\x}
    \pgfmathsetmacro{\dy}{3.25-\y}
    \pgfmathsetmacro{\n}{sqrt(\dx*\dx+\dy*\dy+0.01)}
    \pgfmathsetmacro{\lw}{0.7+0.7*\a}
    \draw[->, cgreen, line width=\lw pt, opacity=\a]
      (\x,\y) -- ++({\dx/\n*0.5},{\dy/\n*0.5});
  }
  \node[font=\small, cgreen] at (2.5,-1.0) {$F^+$ only: fades};

  \draw[->, very thick, black!60] (5.8,1.6) -- (6.8,1.6);

  \begin{scope}[shift={(7.5,0)}]
    \node[font=\sffamily\small, cblue] at (2.5,4.4) {DPO};
    \draw[thick, black!50] (0,0) -- (5,0) -- (2.5,3.25) -- cycle;
    \node at (2.5,3.7) {$y_w$};
    \node at (-0.4,-0.4) {$y_{l_1}$};
    \node at (5.4,-0.4) {$y_{l_2}$};

    \foreach \x/\y in {2.5/1.8, 2.2/2.2, 2.8/2.2} {
      \pgfmathsetmacro{\dx}{2.5-\x}
      \pgfmathsetmacro{\dy}{3.25-\y}
      \pgfmathsetmacro{\n}{sqrt(\dx*\dx+\dy*\dy+0.01)}
      \pgfmathsetmacro{\lw}{0.7+0.7*0.8}
      \draw[->, cblue, line width=\lw pt, opacity=0.8]
        (\x,\y) -- ++({\dx/\n*0.5},{\dy/\n*0.5});
    }
    \foreach \x/\y/\a in {0.9/0.4/0.9, 4.1/0.4/0.9, 2.5/0.5/0.85, 1.3/1.0/0.8, 3.7/1.0/0.8, 1.7/1.5/0.7, 3.3/1.5/0.7} {
      \pgfmathsetmacro{\dlx}{\x-0}
      \pgfmathsetmacro{\drx}{\x-5}
      \pgfmathsetmacro{\useleft}{ifthenelse(\x<2.5,1,0)}
      \pgfmathsetmacro{\dx}{ifthenelse(\x<2.5,\x,\x-5)}
      \pgfmathsetmacro{\dy}{\y}
      \pgfmathsetmacro{\n}{sqrt(\dx*\dx+\dy*\dy+0.01)}
      \pgfmathsetmacro{\lw}{0.7+0.7*\a}
      \draw[->, cred, line width=\lw pt, opacity=\a]
        (\x,\y) -- ++({\dx/\n*0.5},{\dy/\n*0.5});
    }
    \node[font=\small, cblue] at (2.5,-1.0) {$F^+\!-\!F^-$: sustained};
  \end{scope}

  \draw[->, cblue, line width=1.2pt] (3.5,-1.7) -- (4.2,-1.7) node[font=\small, right, black] {$F^+$};
  \draw[->, cred, line width=1.2pt] (8,-1.7) -- (8.7,-1.7) node[font=\small, right, black] {$F^-$};
\end{scope}

\begin{scope}[shift={(17,0)}]
  \node[font=\sffamily\bfseries\small] at (4,5.5) {(b) Axis 2: Margin};

  \draw[->, black!50, thick] (0,0) -- (8,0) node[right] {$t$};
  \draw[->, black!50, thick] (0,0) -- (0,4.8) node[left] {$\Delta$};

  \draw[cblue, line width=1.6pt, smooth, domain=0.05:7.5, samples=50]
    plot (\x, {1.4*ln(1+2*\x)});
  \node[font=\sffamily\small, cblue] at (5.6,4.2) {DPO};

  \draw[cred, line width=1.6pt, smooth, domain=0:7.5, samples=50]
    plot (\x, {1.2*(1-exp(-1.5*\x))});
  \draw[black!30, dashed] (0,1.2) -- (7.8,1.2);
  \node[black!50] at (-0.8,1.2) {$\tfrac{1}{2\tau}$};
  \node[font=\sffamily\small, cred] at (6.6,1.6) {IPO};

  \draw[cpurp, line width=1.2pt, smooth, domain=0:7.5, samples=50, opacity=0.5]
    plot (\x, {2.3*(1-exp(-0.3*\x))});
  \node[font=\sffamily\small, cpurp, opacity=0.6] at (6.9,2.5) {PPO};

  \node[font=\small, black!60, align=center] at (4,-0.9) {sigmoid $\to$ unbounded\\linear $\to$ bounded};
\end{scope}

\begin{scope}[shift={(28,0)}]
  \node[font=\sffamily\bfseries\small] at (4,5.5) {(c) Axis 3: Reference};

  \draw[->, black!50, thick] (0,0) -- (8,0) node[right] {$t$};
  \draw[->, black!50, thick] (0,0) -- (0,4.8) node[left] {$\|\nabla\|$};

  \draw[cblue, line width=1.6pt, smooth, domain=0.05:7.5, samples=50]
    plot (\x, {3.8*exp(-0.6*\x)+0.15});
  \node[font=\sffamily\small, cblue] at (6,0.8) {Frozen};

  \draw[cpurp, line width=1.6pt, smooth, domain=0:7.5, samples=60]
    plot (\x, {1.3 + 0.35*\x + 0.3*sin(\x*2.0 r)*exp(-0.05*\x)});
  \node[font=\sffamily\small, cpurp] at (5.5,4.2) {None};

  \draw[cred, line width=1.2pt, smooth, domain=0:7.5, samples=50, opacity=0.5]
    plot (\x, {2.0 + 0.2*sin(\x*1.3 r)*exp(-0.25*\x)});
  \node[font=\sffamily\small, cred, opacity=0.6] at (6.5,2.45) {Adaptive};

  \node[font=\small, black!60, align=center] at (4,-0.9) {frozen $\to$ stale\\none $\to$ drifts};
\end{scope}

\end{tikzpicture}
\caption{Three axes of the margin dynamics framework. \textbf{(a)}~SFT only attracts toward $y_w$ ($F^+$), with pressure vanishing as $\pi(y_l) \to 0$. DPO adds targeted repulsion $F^-$, sustaining $\Omega(\pi)$ pressure. \textbf{(b)}~Sigmoid weighting (DPO) yields unbounded margin growth; linear weighting (IPO, PPO) converges. \textbf{(c)}~Frozen references cause gradient decay; reference-free methods risk drift; adaptive references stay stable.}
\label{fig:margin_dynamics}
\end{figure}

The most fundamental distinction among methods is how they decrease probability on undesired completions. SFT relies solely on the normalization constraint; preference and RL methods explicitly push probability down.

\begin{definition}[Repulsion types]
\label{def:repulsion}
A method has \emph{targeted repulsion} if its loss explicitly decreases $\pi(y)$ for specific completions. It has \emph{implicit-only repulsion} if the only mechanism is normalization.
\end{definition}

On the two-completion simplex under softmax parameterization, these two repulsion types produce different convergence rates (Figure~\ref{fig:margin_dynamics}a), establishing the first axis of our taxonomy.

\begin{theorem}[Convergence rate separation]
\label{thm:convergence_rate}
On a two-completion simplex under softmax parameterization with learning rate $\eta$:
\emph{(i)}~SFT gives $\dot{\pi}(y_l) = -2\eta\pi(y_l)^2(1-\pi(y_l))$, so $|\dot{\pi}(y_l)| = O(\pi(y_l)^2)$ as $\pi(y_l) \to 0$ (self-limiting).
\emph{(ii)}~DPO gives $\dot{\pi}(y_l) = -2\eta\beta\sigma(-\beta\Delta)\pi(y_l)(1-\pi(y_l))$ with $\sigma(-\beta\Delta) \geq c(\beta, \pi_{\text{ref}}, \delta) > 0$ for $\pi(y_l) \geq \delta > 0$, so $|\dot{\pi}(y_l)| = \Omega(\pi(y_l))$ for $\pi(y_l) \geq \delta$ (sustained while $\pi(y_l)$ remains bounded away from zero). Here $c$ depends on $\beta$, $\pi_{\text{ref}}$, and $\delta$ through the bounded margin $\Delta \leq \log((1-\delta)/\delta) + |\log(\pi_{\text{ref}}(y_w)/\pi_{\text{ref}}(y_l))|$.
\end{theorem}

The $O(\pi^2)$ vs.\ $\Omega(\pi)$ separation means SFT's gradient on bad completions vanishes quadratically while DPO maintains linear pressure. This separates SFT-family from preference methods but does not differentiate \emph{within} the preference family. To do so, we turn to the weighting function governing margin evolution.

\begin{wrapfigure}{r}{0.45\columnwidth}
\vspace{-12pt}
\centering
\begin{tikzpicture}[every node/.style={font=\usefont{T1}{cmss}{m}{n}}, scale=0.72, transform shape]

\definecolor{sblue}{HTML}{4878D0}
\definecolor{sred}{HTML}{D65F5F}
\definecolor{sorange}{HTML}{EE854A}

\begin{axis}[
  width=0.45\columnwidth,
  height=0.32\columnwidth,
  xlabel={\scriptsize Margin $\Delta$},
  ylabel={\scriptsize $V(\Delta)$},
  xmin=-2, xmax=7,
  ymin=-0.3, ymax=5.5,
  axis lines=left,
  every axis x label/.style={at={(ticklabel* cs:1.0)}, anchor=north west},
  every axis y label/.style={at={(ticklabel* cs:1.0)}, anchor=south east},
  xtick=\empty, ytick=\empty,
  clip mode=individual,
  every tick/.style={draw=none},
  axis line style={black!50, line width=0.8pt},
  legend style={
    at={(1.02,0.5)}, anchor=west,
    draw=black!30, line width=0.4pt,
    fill=white, fill opacity=0.92,
    font=\tiny,
    row sep=1pt,
    cells={anchor=west},
  },
]

\addplot[sblue, line width=1.4pt, dashed, domain=-2:6.8, samples=100]
  {1.5*ln(1+exp(-x))};
\addlegendentry{SimPO}

\addplot[sred, line width=1.4pt, solid, domain=-2:6.8, samples=100]
  {0.2*(sqrt(1 + ((x-2)/0.5)^2) - 1) + 1.0};
\addlegendentry{TPO-H}

\addplot[sorange, line width=1.4pt, solid, domain=-2:6.8, samples=100]
  {1.5*(0.85*ln(1+exp(-x)) + 0.15*ln(1+exp(x))) + 2.2};
\addlegendentry{TPO-S}

\addplot[only marks, mark=*, mark size=2.5pt, sred, forget plot]
  coordinates {(2, 1.0)};
\addplot[only marks, mark=*, mark size=2.5pt, sorange, forget plot]
  coordinates {(1.74, 2.83)};

\node[sred, font=\usefont{T1}{cmss}{bx}{n}\tiny,
  fill=white, fill opacity=0.9, text opacity=1,
  inner sep=1.5pt, anchor=north west]
  at (axis cs:2.15,0.9) {$\Delta^*_H$};

\node[sorange, font=\usefont{T1}{cmss}{bx}{n}\tiny,
  fill=white, fill opacity=0.9, text opacity=1,
  inner sep=1.5pt, anchor=south west]
  at (axis cs:1.9,2.9) {$\Delta^*_S$};

\draw[->, sblue, line width=1.1pt] (axis cs:4.5,0.025) -- (axis cs:6.7,0.002);
\node[sblue, font=\usefont{T1}{cmss}{m}{n}\tiny,
  fill=white, fill opacity=0.9, text opacity=1,
  inner sep=1pt, anchor=south]
  at (axis cs:5.6,0.08) {$\Delta\!\to\!\infty$};

\end{axis}
\end{tikzpicture}
\caption{Effective potential $V(\Delta)$ (schematic). SimPO (dashed) has no minimum: margins diverge. TPO-H and TPO-S have finite attractors at $\Delta^*$.}
\label{fig:potential}
\vspace{-10pt}
\end{wrapfigure}
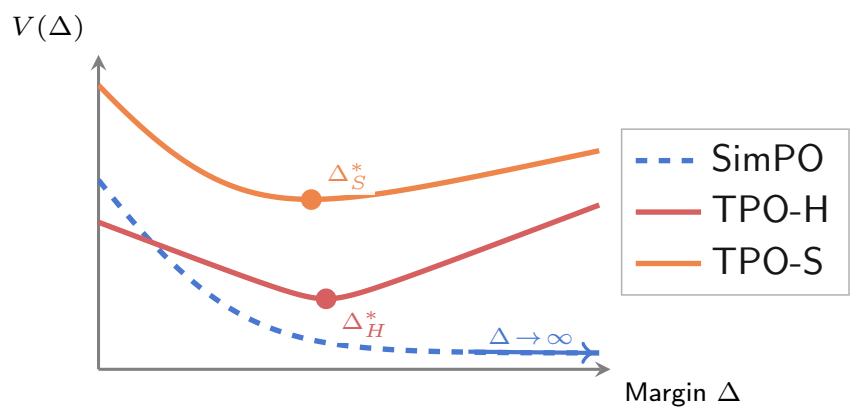

All pairwise preference methods induce $F[\pi](y) \propto w(\Delta) \cdot (\boldsymbol{1}_{y = y_w} - \boldsymbol{1}_{y = y_l})$, differing only in the weighting function $w$: DPO/SimPO use $w(\Delta) = \beta\sigma(-\beta\Delta)$ (sigmoid, saturating); IPO uses $w(\Delta) = \tau - 2\tau^2\Delta$ (linear); KTO uses a bell-shaped $w$ that peaks near $\Delta = 0$ and decays for both large positive and large negative margins, arising from its prospect-theoretic value function \citep{ethayarajh2024kto}; PPO uses $w \propto \Delta^* - \Delta$ (linear, reward-driven). The margin ODE $\dot\Delta = w(\Delta)$ (absorbing $2\eta$ into the time scale; Appendix~\ref{app:proofs}) is a 1D autonomous ODE whose solutions we derive in closed form. All results are stated under gradient flow (continuous time); discretization from finite learning rate introduces $O(\eta)$ errors that do not affect the qualitative predictions (unbounded vs.\ bounded margins, phase transitions) but may shift quantitative fixed points. In practice, Adam's preconditioner acts as a state-dependent time reparameterization of the margin ODE (Appendix~\ref{app:beta_eff_proof}), preserving each trajectory's functional form while the time scale is absorbed into $\beta_{\text{eff}}$.

\begin{theorem}[Margin dynamics without noise]
\label{thm:margin}
\emph{(i)~DPO:} The margin grows without bound: implicit solution $u + \log u = \beta^2 t + C$ where $u = e^{\beta\Delta}$; asymptotically $\Delta(t) \sim \frac{1}{\beta}\log(\beta^2 t)$.
\emph{(ii)~IPO:} The margin converges exponentially: $\Delta(t) = \frac{1}{2\tau} + (\Delta(0) - \frac{1}{2\tau})e^{-2\tau^2 t}$.
\end{theorem}

This distinction sharpens dramatically under noise, revealing a phase transition that constitutes one of our central findings.

\begin{theorem}[Margin dynamics with label noise]
\label{thm:noise}
Under the mean-field approximation (averaging over noise realizations; exact in the large-sample limit), with fraction $\varepsilon \in (0, 1/2)$ of labels flipped:
\emph{(i)~DPO:} unique stable fixed point $\Delta^*_{\text{DPO}} = \frac{1}{\beta}\log\frac{1-\varepsilon}{\varepsilon}$. Without noise, $\Delta^* = \infty$; noise acts as implicit regularization.
\emph{(ii)~IPO:} fixed point $\Delta^*_{\text{IPO}} = \frac{1-2\varepsilon}{2\tau}$ with convergence rate $2\tau^2$ (unchanged by noise).
\end{theorem}

Noise changes DPO's behavior \emph{qualitatively} (unbounded $\to$ bounded) but IPO's only \emph{quantitatively} (shifted fixed point, same rate). This phase transition explains IPO's empirical noise robustness \citep{azar2024general} and directly motivates TPO's label-smoothing term later in this section. The mean-field approximation introduces $O(1/\sqrt{B})$ fluctuations for batch size $B$; with our effective batch size of 4, the transition is smoothed but the fixed-point location is preserved, as confirmed by the consistent $\beta_{\text{eff}}$ across noise levels (Section~\ref{sec:noise_pred}).

The third axis concerns what happens when $N$ pairs share parameters: the gradient becomes dominated by hard examples. Let $v_j = \nabla_\theta\Delta_j$ and $w_i = \nabla_\theta\Delta_{\gamma,i}$ for asymmetric DPO with $\gamma \in [0,1]$.

\begin{theorem}[Gradient concentration]
\label{thm:overshoot}
Under gradient flow, the margin of example $j$ evolves as $\dot{\Delta}_j = \frac{\beta}{N}\sum_{i=1}^N \sigma(-\beta\Delta_{\gamma,i}) \cdot v_j^\top w_i$.
For well-learned examples (i.e., $\pi_\theta(y_l^j) < \pi_{\mathrm{ref}}(y_l^j)$, equivalently $\Delta_{\gamma,j} \gg 1/\beta$), the self-reinforcement term is exponentially small. The dynamics are dominated by hard examples ($\Delta_{\gamma,i} = O(1/\beta)$). Higher $\gamma$ accelerates this concentration.
\end{theorem}

When the concentrated gradient interferes negatively with easy examples ($v_j^\top w_i < 0$), their margins \emph{decrease}, causing margin collapse. This motivates TPO's bounded weighting (Section~\ref{sec:method}).

\paragraph{Bridging theory to practice.} The theorems describe isolated two-completion dynamics, but real training involves $N$ pairs sharing parameters. The gradient concentration result (Theorem~\ref{thm:overshoot}) provides the bridge: each pair's effective dynamics differ from the isolated prediction by cross-example interference terms, formalized through an effective coupling parameter.

\begin{remark}[Effective coupling approximation]
\label{prop:beta_eff}
Consider $N$ training pairs sharing parameters, with margin gradients $v_j = \nabla_\theta\Delta_j$ and loss gradients $w_i = \nabla_\theta\Delta_{\gamma,i}$. The margin of pair $j$ evolves as $\dot{\Delta}_j = (\beta/N)\sum_i \sigma(-\beta\Delta_i) \cdot v_j^\top w_i$ (Theorem~\ref{thm:overshoot}). Define the effective coupling $\beta_{\text{eff},j} = \dot{\Delta}_j / \sigma(-\beta\Delta_j)$. Then, for $\gamma = 1$ (so that $w_j = v_j$):
\begin{equation}
\beta_{\text{eff},j} = \frac{\beta}{N}\left[\|v_j\|^2 + \sum_{i \neq j} \frac{\sigma(-\beta\Delta_i)}{\sigma(-\beta\Delta_j)} \cdot v_j^\top w_i\right],
\end{equation}
where the first term is the self-reinforcement and the sum captures cross-example interference. \textbf{Assumption:} if the cross-example gradient inner products $\{v_j^\top w_i\}_{i \neq j}$ are approximately independent with mean zero and variance $\sigma_\perp^2$ (e.g., when training examples are semantically diverse and gradients do not share a low-rank subspace), then $\beta_{\text{eff}} \approx \beta\|v_j\|^2/N \pm O(\beta\sigma_\perp\sqrt{N_{\text{active}}/N})$, where $N_{\text{active}} = |\{i : \sigma(-\beta\Delta_i) > 1/e\}|$ counts examples with non-negligible gradient weight. The independence assumption is strong and may be violated in practice (e.g., LoRA gradients share a low-rank subspace; thematically correlated datasets induce gradient correlations); the approximation should be interpreted as a scaling relationship rather than a precise prediction. This remark is the principal bridge between the idealized two-completion ODE and multi-example training; its informality reflects the strong independence assumption required, which is satisfied approximately but not rigorously in LoRA training.
\end{remark}

Under the independence assumption, $\beta_{\text{eff}}/\beta$ increases with dataset size: as $N$ grows, more examples reach large margins ($\sigma(-\beta\Delta_i) \to 0$), reducing $N_{\text{active}}$ relative to $N$. The empirical values (${\sim}1/4.4$ to ${\sim}1/1.9$) are post-hoc fits; the remark provides a structural decomposition explaining the scaling trend. The proof is in Appendix~\ref{app:proofs}. In practice, we fit a single $\beta_{\text{eff}}$ per data scale from the clean-data margin trajectory and hold it fixed across all noise conditions; the consistency of this fit across noise levels (Section~\ref{sec:noise_pred}) provides the critical test of the theory.

The two-completion projection is an effective theory: it predicts \emph{functional forms} (logarithmic growth, exponential convergence, phase transitions) that hold across all tested $n \in \{2, 4, 8, 16\}$ completions (Table~\ref{tab:n_completion}), while a single fitted calibration parameter $\beta_{\text{eff}}$ absorbs the gap between the idealized two-completion dynamics and full-scale training (Section~\ref{sec:noise_pred}). We discuss its scope and limitations in Section~\ref{sec:conclusion}. For iterative training, coverage gaps decay as $\Delta_T \leq \Delta_0(1-p_{\min})^{KT}$ (Theorem~\ref{thm:coverage}, Appendix~\ref{app:proofs}). Additional results (PPO dynamics, GRPO gradient vanishing, reference staleness, optimal adaptive reference, and diversity selection) are in Appendix~\ref{app:proofs}.

\begin{table}[t]
\centering
\small
\setlength{\tabcolsep}{3pt}
\begin{tabular}{lccc}
\toprule
Method & Axis 1: Repulsion & Axis 2: Weighting & Axis 3: Reference \\
\midrule
SFT / RFT / STaR & Implicit only & --- & --- \\
PPO / RLHF & Targeted & Linear (non-sat.) & Frozen $\pi_{\text{ref}}$ (KL) \\
GRPO & Targeted (imbal.) & Advantage-weighted & None \\
DPO & Targeted & Sigmoid (saturating) & Frozen $\pi_{\text{ref}}$ \\
IPO & Targeted & Linear (non-sat.) & Frozen $\pi_{\text{ref}}$ \\
SimPO & Targeted & Sigmoid (saturating) & None \\
KTO & Targeted & Bell-shaped (robust) & Implicit (KL term) \\
\textbf{TPO-H} & Targeted & Huber (bounded) & Reference-free \\
\textbf{TPO-S} & Targeted & Sigmoid + smoothing & Reference-free \\
\bottomrule
\end{tabular}
\caption{Three-axis taxonomy of LLM training methods.}
\label{tab:taxonomy}
\end{table}

Table~\ref{tab:taxonomy} organizes the landscape along the three axes, revealing two families: \emph{margin-driven} methods (DPO, IPO, SimPO, KTO) and \emph{reward-driven} methods (PPO, GRPO). PPO bridges both: its KL penalty induces linear margin dynamics like IPO (Proposition~\ref{prop:ppo}). Table~\ref{tab:predictions} maps each theoretical result to its experimental test. Having identified the failure modes along each axis, we derive a loss family that addresses all three simultaneously.

\subsection{From Theory to Loss Design}
\label{sec:method}

The framework is not only explanatory but constructive: specify desired dynamical properties of $w(\Delta)$, then identify losses that produce them. We require five properties: \textbf{D1}~a finite attractor ($w(\Delta^*) = 0$); \textbf{D2}~convergence rate bounded away from zero ($|w|$ bounded below by a positive constant far from $\Delta^*$); \textbf{D3}~stability ($w'(\Delta^*) < 0$); \textbf{D4}~bounded gradient ($|w| \leq c$), preventing the concentration of Theorem~\ref{thm:overshoot}; and \textbf{D5}~asymmetry (stronger push below the target). SimPO's $w(\Delta) = \beta\sigma(-\Delta)$ is strictly positive (\textbf{D1} fails); IPO's $w(\Delta) = \tau - 2\tau^2\Delta$ satisfies \textbf{D1}--\textbf{D3} but grows without bound (\textbf{D4} fails).

We present two complementary instantiations; the ``TPO'' prefix denotes \emph{targeted preference optimization}: both losses are derived from desiderata D1--D5 but via different weighting functions (Huber vs.\ label-smoothed sigmoid), not a shared functional form. \emph{TPO-H} (Huber variant) uses the clamped linear weighting $w^*(\Delta) = \mathrm{clamp}(\Delta^* - \Delta, -c, c)$, satisfying D1--D4 (but not D5). Integrating yields the Huber loss $\cL_{\text{TPO-H}}$ (Appendix~\ref{app:tpo_derivation}). \emph{TPO-S} (label-smoothed sigmoid) exploits Theorem~\ref{thm:noise} directly: adding $\varepsilon > 0$ to SimPO yields $w(\Delta) = (1{-}\varepsilon)\sigma(-\Delta) - \varepsilon\sigma(\Delta)$, with a unique zero at $\Delta^* = \beta^{-1}\log((1{-}\varepsilon)/\varepsilon)$:
\begin{equation}
\label{eq:tpo_s}
\cL_{\text{TPO-S}} = (1-\varepsilon)\bigl(-\log\sigma(\delta)\bigr) + \varepsilon\bigl(-\log\sigma(-\delta)\bigr), \quad \delta = \beta(s_w - s_l - \mu),
\end{equation}
where $s_w, s_l$ are length-normalized log-probabilities (as in SimPO), with $\beta{=}0.1$, $\mu{=}0.5$, $\varepsilon{=}0.05$. (We use $\mu$ for SimPO's target-margin parameter to avoid conflict with the asymmetric weighting $\gamma \in [0,1]$ in $\mathrm{DPO}_\gamma$.) TPO-S satisfies all five desiderata and preserves SimPO's sigmoid curriculum while the label-smoothing term creates a restoring force at $\Delta^*$ (Figure~\ref{fig:potential}). The desiderata identify \emph{when} and \emph{why} these loss designs work, and predict the exact attractor location as a closed-form function of $\varepsilon$. While cDPO \citep{mitchell2023cdpo} also uses label smoothing, it retains a frozen reference and treats $\varepsilon$ as a tuning knob; TPO-S is reference-free, includes a target margin $\mu > 0$, and derives $\varepsilon$ from the phase transition (Theorem~\ref{thm:noise}), giving a closed-form relationship between $\varepsilon$ and the attractor location. The empirical gap between TPO-S and cDPO (2--6 points on IFEval, Table~\ref{tab:ifeval}) reflects the reference axis (Axis~3), not $\varepsilon$ itself. Llama~2's preference margin loss \citep{touvron2023llama} also injects a margin term into the DPO sigmoid, using a per-example offset $m(s_w, s_l)$ derived from Likert-scale annotations. However, such a shift leaves $w(\Delta) = \beta\sigma(-\beta(\Delta-m/\beta))$ strictly positive, so the margin still grows without bound, and only the initial condition is shifted. TPO-S's $\varepsilon$ changes the \emph{sign} of $w$ at $\Delta^*$, a qualitatively different modification that produces the finite attractor.

\paragraph{What the theory adds over grid search.}
A practitioner grid-searching $\varepsilon$ over SimPO would eventually find a good value. The theory predicts a qualitative phase transition: \emph{any} $\varepsilon > 0$ creates a finite attractor while $\varepsilon{=}0$ does not, with the exact attractor location $\Delta^* = \beta^{-1}\log((1{-}\varepsilon)/\varepsilon)$ as a closed-form function of $\varepsilon$. Practitioners can directly control the attractor: $\varepsilon{=}0.01$ gives $\Delta^* \approx 46$ (weaker regularization), $\varepsilon{=}0.1$ gives $\Delta^* \approx 22$ (stronger). We set $\varepsilon{=}0.05$ ($\Delta^* \approx 29$) as a moderate default without per-model tuning. A sweep over $\varepsilon \in \{0.01, \ldots, 0.2\}$ confirms the predicted monotonic ordering (Appendix~\ref{app:eps_sweep}).

\paragraph{The desiderata paradox.}
Both TPO-H and TPO-S satisfy D1--D4; TPO-S additionally satisfies D5. The performance gap is small (0.4--0.6 points on IFEval), confirming that the finite-attractor property (D1--D3) is the primary driver. TPO-S's sigmoid provides exponentially small but nonzero gradients everywhere, giving it a slight edge over TPO-H's hard clipping. Full derivation in Appendix~\ref{app:tpo_derivation}.

The four theorems and the fitted $\beta_{\text{eff}}$ approximation yield concrete, falsifiable predictions: specific functional forms, a noise-induced phase transition, and a concentration-driven collapse pattern, all testable with a single fitted calibration parameter per scale.

\section{Experiments}
\label{sec:experiments}

We validate the framework's predictions and test TPO across three models, two data regimes (on-policy and off-policy), and both preference accuracy and generation quality.
We evaluate on MATH \citep{hendrycks2021math} (transition regime: 37\% base solve rate), GSM8K \citep{cobbe2021gsm8k} (high-coverage: $>$70\%), and AIME 2024 \citep{maa2024aime} (low-coverage: 0\%), spanning the coverage spectrum of Theorem~\ref{thm:coverage}. Primary model: Qwen3-8B with LoRA \citep{hu2022lora} ($r{=}64$). Baselines: SFT, RFT, DPO ($\beta{=}0.1$), IPO ($\tau{=}0.1$), SimPO, KTO, cDPO, ORPO, GRPO; all use default hyperparameters from their respective papers unless noted. Dynamical predictions track margin $\Delta$ over training steps; we fit a single $\beta_{\text{eff}}$ per data scale and use it across all noise conditions without refitting.

\begin{table}[t]
\centering
\small
\setlength{\tabcolsep}{3pt}
\begin{tabular}{llll}
\toprule
Result & Prediction & Experiment & Confirmed? \\
\midrule
\multicolumn{4}{l}{\emph{Main predictions (closed-form margin ODEs)}} \\
Thm.~\ref{thm:margin} & DPO $\Delta \sim \log t$; IPO $\Delta \to 1/(2\tau)$ & Fig.~\ref{fig:predictions} & \cmark~functional form \\
Thm.~\ref{thm:noise} & Noise: DPO phase transition, IPO rate-preserving & Fig.~\ref{fig:noise} & \cmark~8\% error \\
Thm.~\ref{thm:convergence_rate} & $\gamma=0$ vs $\gamma>0$ sharp separation & Fig.~\ref{fig:gamma_drift} & \cmark~qualitative \\
\midrule
\multicolumn{4}{l}{\emph{Supporting predictions (Appendix)}} \\
Thm.~\ref{thm:overshoot} & High $\gamma$ $\to$ collapse & Fig.~\ref{fig:gamma_drift} & \cmark~non-monotonic \\
Thm.~\ref{thm:coverage} & Exponential gap decay & App.~\ref{app:extended_experiments} & \cmark~quantitative \\
Prop.~\ref{prop:grpo} & GRPO fails at low $p$ & App.~\ref{app:extended_experiments} & \cmark~qualitative \\
Prop.~\ref{thm:staleness} & Frozen ref constrains drift & App.~\ref{app:extended_experiments} & \cmark~82$\times$ gap \\
\bottomrule
\end{tabular}
\caption{Theory-to-experiment mapping. Main predictions test the closed-form margin ODEs; supporting predictions are in the appendix. A single calibration parameter $\beta_{\text{eff}}$ maps the two-completion theory to full-scale training.}
\label{tab:predictions}
\end{table}

\subsection{Prospective Predictions}
\label{sec:prospective}

\begin{figure}[t]
\centering
\includegraphics[width=0.78\linewidth]{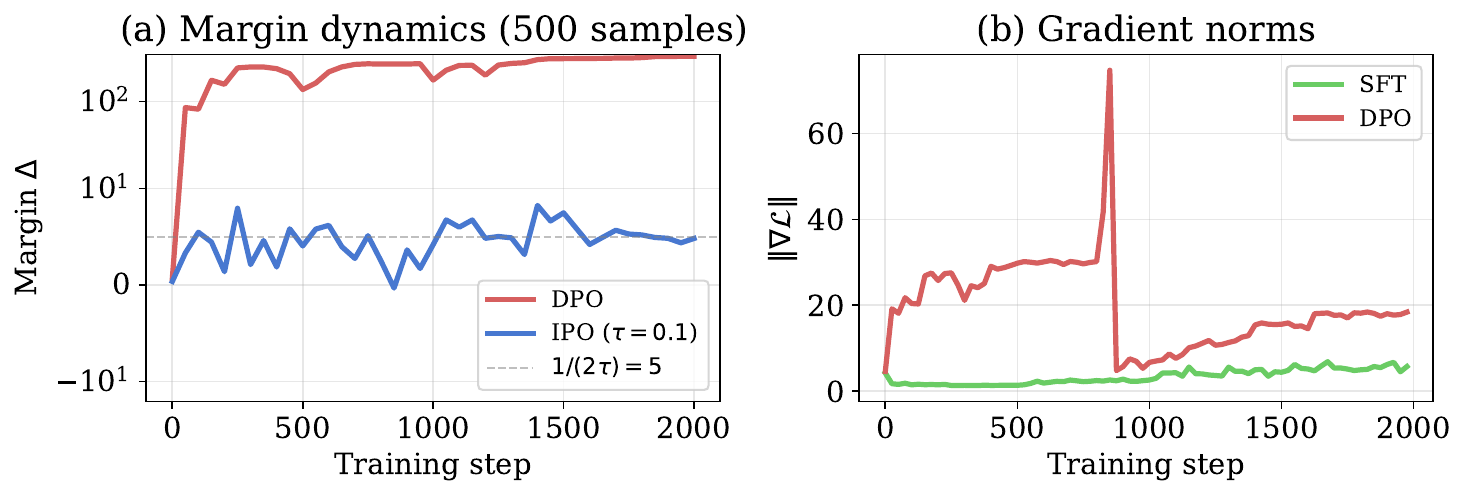}
\caption{Margin and gradient dynamics over 2000 training steps (Theorem~\ref{thm:margin}). Left: DPO's margin grows without bound, consistent with $u + \log u = at + c$ ($R^2 = 0.85$; the remaining variance reflects discrete-time effects and cross-example interference not captured by the isolated ODE). Right: DPO's gradient norm is $10\times$ SFT's, confirming that targeted repulsion sustains $\Omega(\pi)$ gradient pressure (Definition~\ref{def:repulsion}).}
\label{fig:predictions}
\end{figure}

We test three predictions derived from the margin ODEs. The first concerns the functional form of margin trajectories (Theorem~\ref{thm:margin}): DPO's sigmoid weighting predicts unbounded growth as $O(\log t)$, while IPO's linear weighting predicts exponential convergence to $1/(2\tau)$. Figure~\ref{fig:predictions} confirms this separation: DPO reaches 1540 at step 2000 while IPO oscillates near the predicted fixed point of 5. The separation replicates on full MATH (DPO 465 vs.\ IPO $\sim$0), GSM8K (257 vs.\ 8.5), and MBPP \citep{austin2021mbpp} (160 vs.\ 1.1). A $\tau$-sweep validates the IPO fixed point quantitatively across $\tau \in \{0.01, \ldots, 0.2\}$ (Appendix~\ref{app:tau_sweep}).

\begin{figure}[t]
\centering
\begin{minipage}[t]{0.48\linewidth}
\centering
\includegraphics[height=110pt,keepaspectratio]{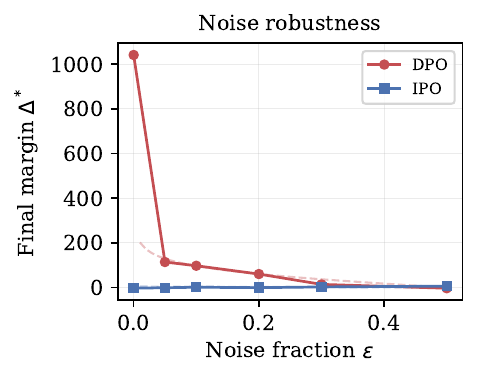}
\caption{Noise phase transition (Theorem~\ref{thm:noise}). DPO's final margin (y-axis) plotted against label-noise fraction $\varepsilon$ (x-axis) collapses from unbounded growth at $\varepsilon{=}0$ to a finite fixed point $\Delta^* = \beta_{\text{eff}}^{-1}\log((1{-}\varepsilon)/\varepsilon)$ for $\varepsilon > 0$ (dashed: theory, solid: empirical). IPO's fixed point shifts with $\varepsilon$ but the convergence rate is preserved.}
\label{fig:noise}
\end{minipage}\hfill
\begin{minipage}[t]{0.48\linewidth}
\centering
\includegraphics[height=110pt,keepaspectratio,clip,trim=0 0 230bp 0]{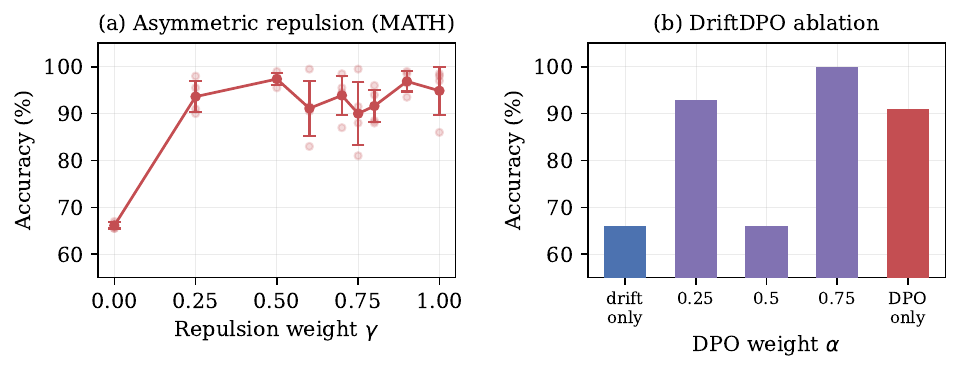}
\caption{Repulsion strength (Theorem~\ref{thm:convergence_rate}). $\gamma{=}0$ plateaus at 66\%; $\gamma{=}0.5$ achieves 97\%; $\gamma{=}0.7$--$0.8$ shows a collapse valley (Theorem~\ref{thm:overshoot}).}
\label{fig:gamma_drift}
\end{minipage}
\end{figure}

This distinction sharpens under noise, providing the second and most striking test.
\label{sec:noise_pred}
Theorem~\ref{thm:noise} predicts that label noise $\varepsilon$ should bound DPO's margin at $\Delta^* = \beta^{-1}\log\frac{1-\varepsilon}{\varepsilon}$ while leaving IPO's convergence rate $2\tau^2$ unchanged. Figure~\ref{fig:noise} confirms: DPO's margin decreases monotonically with noise ($1041 \to 114 \to 97 \to 64$ for $\varepsilon \in \{0, 0.05, 0.1, 0.2\}$), while IPO remains stable. We fit $\beta_{\text{eff}} = \beta/4.4$ once from the clean-data trajectory and hold it constant across all noise conditions; consistency across noise levels ($\beta/4.5$ and $\beta/4.4$ from $\varepsilon{=}0.1$ and $0.2$) confirms the log-formula. The ratio $\beta_{\text{eff}}/\beta$ varies from ${\sim}1/4.4$ (500 samples) to ${\sim}1/1.9$ (full scale); predicting $\beta_{\text{eff}}$ a priori remains open. The qualitative predictions (functional forms, phase transitions, ordering of attractors) are parameter-free; only the quantitative mapping to specific margin values requires $\beta_{\text{eff}}$.

The third axis tests repulsion strength (Theorem~\ref{thm:convergence_rate}): any $\gamma > 0$ should maintain $\Omega(\pi)$ repulsion while $\gamma{=}0$ reverts to $O(\pi^2)$. Figure~\ref{fig:gamma_drift} confirms: $\gamma{=}0$ plateaus at 66\% while $\gamma{=}0.25$ reaches near-ceiling held-out accuracy ($\sim$99\%). Higher $\gamma$ triggers gradient concentration (Theorem~\ref{thm:overshoot}), producing a non-monotonic pattern; measured on the 50 tracked hard examples of Appendix~\ref{app:gamma_collapse} (mean over seeds), $\gamma{=}0.5$ achieves 97\% while $\gamma{=}0.75$ degrades to 87\% (and $\gamma{=}0.25$ recovers to 91\% on this harder tracked set). The pattern is absent on GSM8K (100\% for all $\gamma > 0$), confirming collapse requires hard examples. Standard DPO ($\gamma{=}1.0$) shows the widest variance (82--98\% over 3 seeds). Cross-model validation on Mistral-7B confirms monotonic degradation with $\gamma$ (98\% $\to$ 88\% $\to$ 83\% for $\gamma = 0.25, 0.75, 1.0$).

All three predictions replicate on GSM8K and MBPP (Appendix~\ref{app:gamma_collapse}--\ref{app:multi_benchmark}), confirming that the dynamics are governed by $w(\Delta)$, not by architecture or domain specifics.

\subsection{Theory-Predicted Configurations across Model Scales}
\label{sec:chat_eval}

Having confirmed the framework's descriptive predictions, we test its constructive power: does the theory-predicted configuration match or exceed existing methods? We evaluate on DeepSeek-R1-0528-Qwen3-8B (8B, 9K on-policy pairs), Qwen3.5-9B (9B, 5.4K pairs), and Ministral-3-8B (8B, 2.8K pairs). All methods use LoRA ($r{=}64$); TPO-S uses $\beta{=}0.1$, $\mu{=}0.5$, $\varepsilon{=}0.05$ (predicted by Theorem~\ref{thm:noise}); TPO-H uses $\beta{=}2.0$, $\mu{=}1.0$. No hyperparameter tuning was performed on $\varepsilon$; the theory predicts that \emph{any} $\varepsilon > 0$ creates a finite attractor. We report IFEval (objective instruction-following) and UltraFeedback preference accuracy.

\begin{table}[t]
\centering
\small
\begin{tabular}{lccc}
\toprule
 & DS-R1-8B & QW3.5-9B & MI-8B \\
\midrule
\textbf{TPO-S} & \textbf{66.8 $\pm$ 0.3} & \textbf{61.4 $\pm$ 0.4} & \textbf{76.5 $\pm$ 0.3} \\
TPO-H & 66.4 $\pm$ 0.3 & 60.8 $\pm$ 0.3 & \textit{76.1 $\pm$ 0.9} \\
SimPO & \textit{66.5 $\pm$ 0.6} & \textit{60.9 $\pm$ 0.2} & 73.9 $\pm$ 1.4 \\
DPO & 65.1 $\pm$ 0.2 & 60.6 $\pm$ 0.8 & 71.9 $\pm$ 0.6 \\
KTO & 64.2 $\pm$ 0.8 & 60.4 $\pm$ 0.5 & 72.2 $\pm$ 1.0 \\
cDPO & 64.1 $\pm$ 0.7 & 59.8 $\pm$ 0.4 & 70.6 $\pm$ 0.5 \\
IPO & 64.2 $\pm$ 0.7 & 59.7 $\pm$ 0.6 & 70.3 $\pm$ 1.4 \\
ORPO & 52.6 $\pm$ 1.7 & 51.2 $\pm$ 7.5 & 49.5 $\pm$ 6.2 \\
SFT & 63.1 & 59.8 & 70.3 \\
\bottomrule
\end{tabular}
\caption{IFEval instruction-following accuracy (\%) on UltraFeedback-trained models (mean $\pm$ std, 5 seeds). IFEval is an objective benchmark with no judge model. The theory-derived TPO-S attains the highest mean among the compared methods on all three, confirming the predicted attractor is a competitive operating point. TPO-S adds negligible wall-clock overhead over SimPO: a single extra $\log\sigma(-\delta)$ evaluation per example (the $\varepsilon$-weighted label-smoothing term in Eq.~\ref{eq:tpo_s}).\protect\footnotemark}
\label{tab:ifeval}
\end{table}

\begin{table}[t]
\centering
\small
\setlength{\tabcolsep}{2.5pt}
\begin{tabular}{l|ccc|ccc}
\toprule
& \multicolumn{3}{c|}{On-policy (Math)} & \multicolumn{3}{c}{Off-policy (UltraFeedback)} \\
& DS-8B & QW-9B & MI-8B & DS-8B & QW-9B & MI-8B \\
\midrule
\textbf{TPO-S} & \textbf{73.6}{\scriptsize$\pm$.5} & \textbf{72.0}{\scriptsize$\pm$1.9} & 59.4{\scriptsize$\pm$1.1} & \textbf{55.2}{\scriptsize$\pm$1.9} & \textbf{59.4}{\scriptsize$\pm$1.9} & \textit{58.2}{\scriptsize$\pm$2.2} \\
SimPO & \textit{72.4}{\scriptsize$\pm$1.9} & \textit{69.4}{\scriptsize$\pm$2.4} & 58.6{\scriptsize$\pm$6.2} & \textit{51.2}{\scriptsize$\pm$.8} & \textit{54.6}{\scriptsize$\pm$2.4} & \textbf{60.2}{\scriptsize$\pm$.8} \\
TPO-H & 71.4{\scriptsize$\pm$2.6} & 68.0{\scriptsize$\pm$1.9} & \textbf{59.8}{\scriptsize$\pm$1.9} & 47.8{\scriptsize$\pm$.8} & 53.4{\scriptsize$\pm$4.2} & 51.2{\scriptsize$\pm$1.1} \\
KTO & 50.2{\scriptsize$\pm$1.3} & 64.2{\scriptsize$\pm$.8} & \textit{59.6}{\scriptsize$\pm$1.1} & 46.8{\scriptsize$\pm$1.1} & 52.0{\scriptsize$\pm$2.0} & 52.8{\scriptsize$\pm$1.5} \\
DPO & 47.0{\scriptsize$\pm$2.0} & 61.0{\scriptsize$\pm$2.3} & 55.6{\scriptsize$\pm$8.8} & 45.4{\scriptsize$\pm$.5} & 47.0{\scriptsize$\pm$4.8} & 55.2{\scriptsize$\pm$2.8} \\
cDPO & 46.6{\scriptsize$\pm$1.5} & 59.0{\scriptsize$\pm$.7} & 57.2{\scriptsize$\pm$1.5} & 45.2{\scriptsize$\pm$.4} & 45.6{\scriptsize$\pm$.9} & 46.0{\scriptsize$\pm$1.2} \\
IPO & 43.8{\scriptsize$\pm$.4} & 55.4{\scriptsize$\pm$1.5} & 50.8{\scriptsize$\pm$.4} & 43.4{\scriptsize$\pm$.5} & 42.8{\scriptsize$\pm$1.3} & 45.2{\scriptsize$\pm$.8} \\
ORPO & 42.5{\scriptsize$\pm$.0} & 53.8{\scriptsize$\pm$3.6} & 46.5{\scriptsize$\pm$.0} & 42.4{\scriptsize$\pm$.8} & 41.8{\scriptsize$\pm$.4} & 43.0{\scriptsize$\pm$.0} \\
\bottomrule
\end{tabular}
\caption{Preference accuracy (\%, mean $\pm$ std, 5 seeds) on on-policy math data and off-policy UltraFeedback \citep{cui2024ultrafeedback} (60K pairs). TPO-S leads or ties on most settings. The reference-free vs.\ reference-dependent gap exceeds 24 points on DS-R1-8B.}
\label{tab:combined_results}
\end{table}
\footnotetext{$p$-values (TPO-S vs.\ SimPO, Welch's $t$-test, 5 seeds): MI $p{=}0.013$, QW $p{=}0.047$, DS $p{=}0.36$.}

The theory-derived TPO-S attains the highest mean IFEval score among the compared methods across all three models (Table~\ref{tab:ifeval}), with the effect significant on MI-8B (+2.6 over SimPO, $p{=}0.013$) and within noise on DS-8B ($p{=}0.36$); we read this as confirming the predicted attractor is a competitive operating point, not as a state-of-the-art claim. Method ranking aligns with the taxonomy: reference-free methods with finite attractors (TPO-S, TPO-H, SimPO) occupy the top three positions on every model, reference-dependent methods form a middle tier, and ORPO \citep{hong2024orpo} falls far behind. GRPO and PPO are online RL methods analyzed theoretically (Proposition~\ref{prop:grpo}); we exclude them from Table~\ref{tab:ifeval} to keep the comparison within the offline-pairwise setting.

\paragraph{Ablation: reference vs.\ attractor.}
Table~\ref{tab:ifeval} implicitly contains a $2{\times}2$ ablation over reference model (present/absent) and finite attractor ($\varepsilon{>}0$ / $\varepsilon{=}0$), with DPO, cDPO, SimPO, and TPO-S filling the four cells.
Removing the reference contributes $+1.2$ points on average (SimPO $-$ DPO: $+1.4$, $+0.3$, $+2.0$), while adding the attractor on top of a reference-free loss contributes $+1.1$ (TPO-S $-$ SimPO: $+0.3$, $+0.5$, $+2.6$).
Adding the attractor to a reference-based loss \emph{hurts}: cDPO $-$ DPO averages $-1.0$, negative on every model—when the frozen reference already bounds margins, additional clipping reduces useful gradient signal.
The two axes are complementary: TPO-S combines both, and neither alone matches its gains on MI-8B ($+4.6$ over DPO).

Table~\ref{tab:combined_results} shows two robust patterns and one suggestive one. Robustly: TPO-S leads or ties on most settings, and reference-free methods consistently outscore reference-dependent ones (gap exceeding 24 points on DS-8B on-policy, Proposition~\ref{thm:staleness}). Suggestively, and within the seed variance of this table, TPO-S retains its edge off-policy on DS-8B and QW-9B, consistent with the finite attractor providing regularization under distribution mismatch; we do not over-read the per-setting margins here given the seed-to-seed spread. Method rankings are preserved under the standard SFT $\to$ preference pipeline (Appendix~\ref{app:simpo_replication}).

Additional experiments (method rankings, diversity/coverage, $n$-completion validation, ablations) are in Appendix~\ref{app:extended_experiments}. On mathematical reasoning, TPO reaches 98.7\% on MATH vs.\ 93.8\% for DPO (Appendix~\ref{app:tpo_math}), with the gain concentrated in the transition regime (Theorem~\ref{thm:overshoot}); on GSM8K both reach 100\%; on AIME both score $\sim$12\% ($p_t \approx 0$, Theorem~\ref{thm:coverage}).

\paragraph{Label-smoothing sensitivity.}
Theorem~\ref{thm:noise} predicts that any $\varepsilon > 0$ creates a finite attractor at $\Delta^* {=} \beta^{-1}\log((1{-}\varepsilon)/\varepsilon)$, with higher $\varepsilon$ yielding a lower equilibrium. A sweep on DS-R1-8B (UltraFeedback, 60K pairs) confirms: final margins decrease monotonically---$1605 \to 1264 \to 832 \to 455 \to 341$ for $\varepsilon \in \{0.01, 0.03, 0.05, 0.1, 0.2\}$---while SimPO ($\varepsilon{=}0$) diverges. Observed margins are $25$--$35\times$ the closed-form $\Delta^*$ (e.g., $832$ vs.\ $29$ at $\varepsilon{=}0.05$), reflecting $\beta_{\text{eff}}$ renormalization (Remark~\ref{prop:beta_eff}). Downstream accuracy is flat across $\varepsilon$ ($49$--$52\%$), confirming that any $\varepsilon > 0$ suffices (Appendix~\ref{app:eps_sweep}).

\paragraph{Cross-Architecture Universality}
\label{sec:crossmodel}

\begin{figure}[t]
\centering
\includegraphics[width=0.85\linewidth]{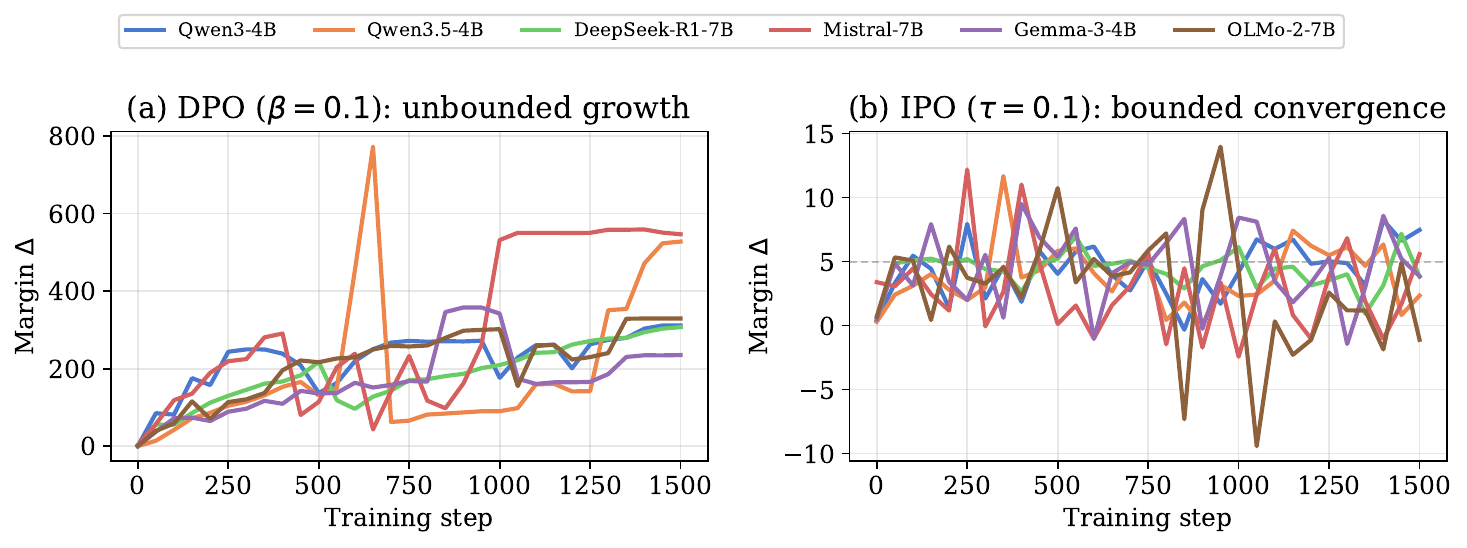}
\caption{Margin dynamics replicated across five model families (Qwen, Gemma, OLMo, Mistral, DeepSeek). DPO's margin grows without bound and IPO's converges near the predicted fixed point $1/(2\tau) = 5$ (dashed line) on every architecture. Only the growth rate varies across families.}
\label{fig:crossmodel}
\end{figure}

The theory operates on the probability simplex, shared by all autoregressive models with softmax outputs, so its predictions should be architecture-agnostic. We replicate Prediction~1 on six additional models spanning five families. In every case, DPO's margin grows to $204$--$629$ while IPO's remains bounded near $1/(2\tau) = 5$ (Figure~\ref{fig:crossmodel}). Growth rates vary $3\times$ across families, reflecting differences in $\beta_{\text{eff}}$, but the functional form is invariant: logarithmic for DPO, exponential convergence for IPO, despite different tokenizers, vocabularies, and parameter counts.

\FloatBarrier
\section{Related Work}
\label{sec:related}

\paragraph{Alignment methods and unified frameworks.}
We analyze RLHF/PPO \citep{ouyang2022rlhf,schulman2017ppo}, DPO \citep{rafailov2023dpo}, IPO \citep{azar2024general}, SimPO \citep{meng2024simpo}, KTO \citep{ethayarajh2024kto}, GRPO \citep{shao2024grpo}, and others as instances of a single dynamical mechanism. \citet{deng2026drifting} introduce the $F^+/F^-$ decomposition for image generation on continuous velocity fields; we borrow this decomposition but work on a discrete simplex with 1D margin ODEs. Several frameworks characterize \emph{what} each method optimizes---$f$-PO \citep{han2025fpo} via $f$-divergences, RPO \citep{liu2025rpo} via a common objective, DCR \citep{li2025dcr} via gradient flow---while we characterize \emph{how} the optimization unfolds. Llama~2's preference margin loss \citep{touvron2023llama} shifts the DPO sigmoid by a per-example offset derived from Likert annotations; in our taxonomy this leaves $w(\Delta)$ strictly positive (no attractor). The noise phase transition of Theorem~\ref{thm:noise} exemplifies the difference: $f$-PO can show that DPO and IPO minimize different divergences, but cannot predict that DPO's margin grows logarithmically while IPO's converges exponentially, or that noise creates a phase transition in one but not the other.

\paragraph{Failure modes and iterative training.}
Concurrent work identifies individual DPO pathologies—gradient imbalance \citep{yan2024threed,ma2025gradient}, likelihood displacement \citep{razin2024likelihood}, noise fragility \citep{chowdhury2024rdpo}, GRPO zero-gradient collapse \citep{yu2025adaptive}—which our framework unifies as instances of a single cross-method pattern governed by the three axes. DAPO \citep{yu2025dapo} removes the reference model entirely (reference-free, bounded convergence in our taxonomy). Iterative methods \citep{yuan2024rft,zelikman2022star,guo2024onlinedpo} refresh the training distribution; our coverage theorem (Theorem~\ref{thm:coverage}) formalizes when they succeed ($p_t > 0$) and stall ($p_t \to 0$).

\section{Conclusion}
\label{sec:conclusion}

We showed that major pairwise LLM alignment methods are instances of a single dynamical mechanism: a vector field on the probability simplex whose margin ODE admits closed-form solutions. The three main predictions (margin trajectories, noise-induced phase transitions, and repulsion strength) are confirmed across five model families. We validated the framework's constructive prediction that \emph{any} $\varepsilon > 0$ label smoothing creates a finite attractor for SimPO across three models; the theory-derived TPO-S ($\varepsilon{=}0.05$) is a competitive operating point, attaining the highest mean IFEval among the compared methods on all three, which confirms the predicted attractor is usable rather than establishing a new state-of-the-art loss.

\paragraph{Limitations.} The two-completion simplex is the central modeling choice: qualitative predictions hold at $n \in \{2, 4, 8, 16\}$ (Table~\ref{tab:n_completion}), but quantitative accuracy requires $\beta_{\text{eff}}$, fitted per experiment ($2.3\times$ variation across scales). The margin is a sufficient state variable only in the leakage-free / independent-gradient regime: when the chosen and rejected gradients are strongly aligned ($\langle g_w, g_l\rangle > \lVert g_w\rVert^2$, with $g=\nabla_\theta\log\pi$), mass escaping to unsampled completions can decouple the margin from $\pi(y_w)$ (likelihood displacement, \citealp{razin2024likelihood}), so the margin captures relative preference dynamics and predicting downstream quality from it alone is out of scope---consistent with the weak margin--accuracy correlation ($\rho \approx 0.13$) on AIME. TPO-S's gain over SimPO is modest (1.2 pts on DS~8B); on MI~8B KTO leads on chat. The $\gamma$-sweep collapse valley shows high seed variance (82--98\%). Experiments use LoRA ($r{=}64$) and Adam; Adam acts as a time reparameterization absorbed into $\beta_{\text{eff}}$ (Appendix~\ref{app:beta_eff_proof}), and the residual $R^2{=}0.85$ (Figure~\ref{fig:predictions}) reflects discrete-time and cross-example effects.

\bibliographystyle{plainnat}

\bibliography{references}

\appendix

\section{Proofs}
\label{app:proofs}

\begin{remark}[Extension beyond two completions]
\label{rem:n_completions}
The two-completion analysis is exact at $n = 2$ (no leakage by construction). As $n \to \infty$ with all completions labeled, repelling every incorrect completion concentrates mass on correct ones, and the analysis again becomes tight, though this limit is unrealistic in practice. The approximation error is largest at intermediate $n$, where repelling a subset of incorrect completions redistributes mass to unsampled ones, the displacement phenomenon of \citet{asadi2026textttcdpo}. This makes the two-completion simplex a \emph{conservative} approximation: it predicts the advantage of targeted repulsion in the leakage-free regime, while the actual advantage with $n > 2$ is strictly larger. We validate empirically that the qualitative predictions (unbounded DPO margins, bounded IPO margins) hold at $n \in \{2, 4, 8, 16\}$ (Appendix~\ref{app:supporting}), though quantitative accuracy at $n > 2$ with the same $\beta_{\text{eff}}$ remains unvalidated. Diversity-aware selection (Proposition~\ref{prop:diversity}) mitigates leakage by covering the incorrect-completion space, consistent with the +29\% GRPO gain at $K=4$.
\end{remark}

\paragraph{Notation and conventions.} All results are stated on a two-completion simplex $\{y_w, y_l\}$ under softmax parameterization $\pi(y_i) = e^{\theta_i}/(e^{\theta_w} + e^{\theta_l})$. We assume gradient flow (continuous time) with full-batch gradients; SGD noise from mini-batching is not analyzed but does not affect the qualitative predictions.

\begin{remark}[Connection to generative modeling]
\label{rem:generative}
The vector field $F$ on the probability simplex is the same mathematical object studied in continuous normalizing flows \citep{lipman2022flow} and flow matching \citep{geng2025meanflow}. In those settings, a learned velocity field transports a source distribution to a target along a probability path; here, the loss function \emph{induces} the velocity field on the margin coordinate $\Delta = \log(\pi(y_w)/\pi(y_l))$, yielding the 1D autonomous ODEs we solve in closed form. The noise fixed point of Theorem~\ref{thm:noise}--where the forward and reverse ``flows'' balance--is the discrete analog of score equilibrium in diffusion models \citep{ho2020denoising,song2020score}: label-smoothed TPO (Section~\ref{sec:method}) exploits this by engineering a stable equilibrium at a prescribed margin, converting the noise phase transition from a failure mode into a design principle.
\end{remark}

For IPO, we analyze the reference-free variant $\cL = (\tau\Delta_\theta - 1/2)^2$ where $\Delta_\theta = \log(\pi(y_w)/\pi(y_l))$; with a frozen reference, the dynamics are identical up to a constant shift $\Delta_{\text{ref}}$ in the fixed point. For DPO, the full margin is $\tilde{\Delta} = \Delta_\theta - \Delta_{\text{ref}}$ where $\Delta_{\text{ref}} = \log(\pi_{\text{ref}}(y_w)/\pi_{\text{ref}}(y_l))$ is constant under a frozen reference; the ODE $d\tilde{\Delta}/dt = \beta\sigma(-\beta\tilde{\Delta})$ is autonomous in $\tilde{\Delta}$, so the reference shifts the initial condition but does not change the functional form of the solution. The key identity is the softmax Jacobian:
\begin{equation}
\label{eq:softmax_jacobian}
\dot{\pi}(y_l) = \pi(y_l)(1-\pi(y_l))\left(\dot{\theta}_l - \dot{\theta}_w\right).
\end{equation}
For any loss $\cL$, gradient descent gives $\dot{\theta}_y = -\eta \frac{\partial \cL}{\partial \theta_y}$. Since DPO and IPO losses depend on $\theta_w - \theta_l$, both logits receive equal and opposite gradients, yielding $\dot{\theta}_w - \dot{\theta}_l = 2\eta \cdot w(\Delta)$ where $w(\Delta)$ is the method-specific weighting. We absorb $2\eta$ into the time scale, writing $\frac{d\Delta}{dt} = w(\Delta)$.

\subsection{Axis 1: Convergence Rate Separation (Theorem~\ref{thm:convergence_rate})}

\emph{Proof sketch.} Compute the softmax gradient for each loss, substitute into the Jacobian~\eqref{eq:softmax_jacobian}, and compare the rate at which $|\dot{\pi}(y_l)|$ vanishes as $\pi(y_l) \to 0$.

\begin{proof}
\emph{Part (i): SFT.} The cross-entropy loss $\cL = -\log\pi(y_w)$ gives gradients $\dot{\theta}_y = \eta(p^*(y) - \pi(y))$. With $p^*(y_w) = 1$, $p^*(y_l) = 0$:
\begin{align}
\dot{\theta}_w = \eta\pi(y_l), \qquad \dot{\theta}_l = -\eta\pi(y_l).
\end{align}
Thus $\dot{\theta}_l - \dot{\theta}_w = -2\eta\pi(y_l)$. By~\eqref{eq:softmax_jacobian}:
\begin{equation}
\dot{\pi}(y_l) = -2\eta\pi(y_l)^2(1-\pi(y_l)) = O(\pi(y_l)^2) \text{ as } \pi(y_l) \to 0.
\end{equation}

\emph{Part (ii): DPO.} The loss $\cL = -\log\sigma(\beta\tilde{\Delta})$ with $\tilde{\Delta} = (\theta_w - \theta_l) - \log(\pi_{\text{ref}}(y_w)/\pi_{\text{ref}}(y_l))$ gives:
\begin{align}
\frac{\partial \cL}{\partial \theta_w} = -\beta\sigma(-\beta\tilde{\Delta}), \qquad
\frac{\partial \cL}{\partial \theta_l} = \beta\sigma(-\beta\tilde{\Delta}).
\end{align}
So $\dot{\theta}_l - \dot{\theta}_w = -2\eta\beta\sigma(-\beta\Delta)$, and by~\eqref{eq:softmax_jacobian}:
\begin{equation}
\dot{\pi}(y_l) = -2\eta\beta\sigma(-\beta\Delta)\pi(y_l)(1-\pi(y_l)).
\end{equation}
When $\pi(y_l) \geq \delta > 0$, the margin $\Delta$ is bounded above, so $\sigma(-\beta\Delta) \geq c(\beta, \pi_{\text{ref}}, \delta) > 0$ where $c$ depends on $\beta$, $\pi_{\text{ref}}$, and $\delta$ through the bounded margin. Therefore $|\dot{\pi}(y_l)| = \Omega(\pi(y_l))$: sustained.
\end{proof}

\subsection{Axis 2: Margin Dynamics (Theorems~\ref{thm:margin}--\ref{thm:noise})}

\paragraph{Theorem~\ref{thm:margin} (without noise).}
\emph{Proof sketch.} Solve the 1D ODE $d\Delta/dt = w(\Delta)$ for each weighting function. DPO's sigmoid yields a nonlinear ODE solvable via substitution; IPO's linear weighting gives a standard linear ODE.

\begin{proof}
\emph{(i) DPO.} The ODE $\frac{d\Delta}{dt} = \frac{\beta}{1 + e^{\beta\Delta}}$ becomes, under $u = e^{\beta\Delta}$:
\begin{equation}
\frac{du}{dt} = \frac{\beta^2 u}{1+u}.
\end{equation}
Separating variables and integrating: $\log u + u = \beta^2 t + C$ where $C = \beta\Delta(0) + e^{\beta\Delta(0)}$. For large $t$, $u$ dominates $\log u$, so $u(t) \sim \beta^2 t$ and $\Delta(t) \sim \frac{1}{\beta}\log(\beta^2 t)$: unbounded but logarithmic. (This asymptotic is valid for $\beta^2 t \gg 1$; at early times the implicit solution should be used directly.)

\emph{(ii) IPO.} The ODE $\frac{d\Delta}{dt} = \tau - 2\tau^2\Delta$ is linear with fixed point $\Delta^* = 1/(2\tau)$. Writing $\delta = \Delta - \Delta^*$:
\begin{equation}
\frac{d\delta}{dt} = -2\tau^2\delta \implies \Delta(t) = \frac{1}{2\tau} + \left(\Delta(0) - \frac{1}{2\tau}\right)e^{-2\tau^2 t}.
\end{equation}
Exponential convergence to $1/(2\tau)$ with rate $2\tau^2$.
\end{proof}

\paragraph{Theorem~\ref{thm:noise} (with label noise).}
\emph{Proof sketch.} Average the ODE over correct and flipped pairs, treating the noise as i.i.d.\ across examples. This is a mean-field approximation: we replace the per-example noise realization with its expectation, which becomes exact in the large-sample limit (law of large numbers applied to the batch gradient). For finite batches, the approximation introduces fluctuations of order $O(1/\sqrt{B})$ where $B$ is the batch size. For DPO, the opposing sigmoid terms create a stable fixed point. For IPO, the linear structure preserves the convergence rate.

\begin{proof}
\emph{(i) DPO.} The averaged ODE is:
\begin{equation}
\frac{d\Delta}{dt} = (1-\varepsilon)\beta\sigma(-\beta\Delta) - \varepsilon\beta\sigma(\beta\Delta).
\end{equation}
Setting to zero and using $\sigma(-x) = e^{-x}\sigma(x)$: $(1-\varepsilon)e^{-\beta\Delta^*} = \varepsilon$, giving $\Delta^* = \frac{1}{\beta}\log\frac{1-\varepsilon}{\varepsilon}$. Stability: differentiating the full averaged field,
$f'(\Delta) = -(1{-}\varepsilon)\beta^2\sigma(\beta\Delta)\sigma(-\beta\Delta) - \varepsilon\beta^2\sigma(\beta\Delta)\sigma(-\beta\Delta) = -\beta^2\sigma(\beta\Delta)\sigma(-\beta\Delta) < 0$ everywhere.

\emph{(ii) IPO.} A flipped pair has effective margin $-\Delta$, giving loss $(\tau\Delta + 1/2)^2$. Averaging:
\begin{equation}
\frac{d\Delta}{dt} = -2\tau^2\Delta + \tau(1-2\varepsilon).
\end{equation}
Fixed point $\Delta^* = (1-2\varepsilon)/(2\tau)$, rate $2\tau^2$ unchanged. Noise shifts the target but not the speed.
\end{proof}

\begin{proposition}[GRPO gradient vanishes with sparse rewards]
\label{prop:grpo}
For GRPO with binary rewards, group size $K$, success probability $p$: $\E[\|\nabla\cL\|^2] \propto p(1-p) \cdot G^2/K$, maximized at $p = 1/2$ and vanishing as $p \to 0$ or $p \to 1$. The zero-gradient probability is $(1-p)^K + p^K$.
\end{proposition}

\begin{proof}
The advantage $A_i = R_i - \bar{R}$ gives $\sum A_i^2 = k(K-k)/K$. Since $\E[k(K-k)] = Kp(1-p)(K-1)$:
\begin{equation}
\E\!\left[\frac{1}{K^2}\sum A_i^2\right] = \frac{(K-1)p(1-p)}{K^2} \sim \frac{p(1-p)}{K}.
\end{equation}
Maximized at $p=1/2$, vanishing as $p \to 0$ or $1$. Zero-gradient probability: $(1-p)^K + p^K$.
\end{proof}

\subsection{Axis 3: Reference Strategy (Proposition~\ref{thm:staleness}, Proposition~\ref{prop:optimal_ref})}

\begin{proposition}[Gradient decay under staleness]
\label{thm:staleness}
For any method with frozen $\pi_{\text{ref}}$ whose gradient includes $\sigma(-\beta\Delta)$, the per-example gradient norm is $\|\nabla\cL\| = \beta \cdot \sigma(-\beta\Delta) \cdot \|\nabla\log\pi(y_w) - \nabla\log\pi(y_l)\|$. As training progresses, $\Delta$ grows on correctly-learned examples, driving $\sigma(-\beta\Delta) \to 0$. Reference-free methods (SimPO, ORPO) avoid this decay but still experience sigmoid-based gradient decay from growing margins; the staleness proposition isolates the \emph{additional} decay caused by the frozen reference.
\end{proposition}

\begin{proof}
The DPO gradient is $\nabla_\theta \cL = -\beta\sigma(-\beta\Delta)[\nabla_\theta\log\pi_\theta(y_w|x) - \nabla_\theta\log\pi_\theta(y_l|x)]$. Taking norms:
\begin{equation}
\|\nabla_\theta \cL\| = \beta \cdot \sigma(-\beta\Delta) \cdot \|\nabla_\theta\log\pi_\theta(y_w|x) - \nabla_\theta\log\pi_\theta(y_l|x)\|.
\end{equation}
Since $\Delta = \Delta_\theta - \Delta_{\text{ref}}$ and $\Delta_\theta$ grows on correctly-learned examples (as established by Theorem~\ref{thm:margin} for DPO), $\sigma(-\beta\Delta) \to 0$. The log-probability gradient term is bounded for finite-parameter models, so the sigmoid decay dominates. For SimPO, $\Delta = \Delta_\theta - \mu$ with fixed $\mu$, so no \emph{reference-staleness} decay occurs. However, SimPO still experiences sigmoid-based gradient decay from growing margins ($\sigma(-\beta\Delta_\theta) \to 0$ as $\Delta_\theta$ grows). The distinction is that DPO has \emph{both} margin-based decay and reference-staleness decay, while SimPO has only the former.
\end{proof}

\begin{proposition}[Optimal adaptive reference]
\label{prop:optimal_ref}
The reference minimizing $\lambda_1 \KL(\pi_{\text{ref}} \| \pi_\theta) + \lambda_2 \KL(\pi_{\text{ref}} \| \pi_0)$ is
$\pi_{\text{ref}}^*(y) \propto \pi_0(y)^{1-\alpha} \pi_\theta(y)^{\alpha}$
with $\alpha = \lambda_1/(\lambda_1 + \lambda_2)$. At $\alpha = 0$: frozen (DPO). At $\alpha = 1$: self-referencing (SimPO-like).
\end{proposition}

\begin{proof}
The objective $J(\pi_{\text{ref}}) = \lambda_1 \KL(\pi_{\text{ref}} \| \pi_\theta) + \lambda_2 \KL(\pi_{\text{ref}} \| \pi_0)$ is strictly convex. Setting the Lagrangian derivative to zero:
\begin{equation}
(\lambda_1 + \lambda_2)\log\pi_{\text{ref}}(y) = \lambda_1\log\pi_\theta(y) + \lambda_2\log\pi_0(y) + \text{const}.
\end{equation}
Exponentiating: $\pi_{\text{ref}}^*(y) \propto \pi_\theta(y)^{\alpha} \cdot \pi_0(y)^{1-\alpha}$ with $\alpha = \lambda_1/(\lambda_1+\lambda_2)$.

\emph{Note:} We minimize $\KL(\pi_{\text{ref}} \| \pi_\theta)$ (forward KL) rather than $\KL(\pi_\theta \| \pi_{\text{ref}})$ (staleness) because the forward direction yields a closed form. Both vanish at $\pi_{\text{ref}} = \pi_\theta$.
\end{proof}

\subsection{Diversity and PPO (Propositions~\ref{prop:diversity}--\ref{prop:ppo})}

\begin{proposition}[Diversity improves field estimation]
\label{prop:diversity}
For a weighted estimate $\hat{V} = \sum w_i v_i$ with independent $v_i$, $\E[v_i] = V$, $\Var(v_i) = \sigma^2 I$, and $\sum w_i = 1$: $\text{MSE}(\hat{V}) = \sigma^2/d_{\text{eff}}$ where $d_{\text{eff}} = (\sum w_i^2)^{-1} \leq N$, with equality iff weights are uniform.
\end{proposition}

\begin{proof}
For $\hat{V} = \sum w_i v_i$ with $\sum w_i = 1$ and independent $v_i$ with $\E[v_i]=V$, $\Var(v_i)=\sigma^2 I$:
\begin{equation}
\text{MSE}(\hat{V}) = \Var(\hat{V}) = \sigma^2 \sum w_i^2 = \sigma^2/d_{\text{eff}}.
\end{equation}
By Cauchy--Schwarz, $1 = (\sum w_i)^2 \leq N\sum w_i^2$, so $d_{\text{eff}} \leq N$ with equality iff $w_i = 1/N$.
\end{proof}

\begin{proposition}[PPO-KL margin dynamics]
\label{prop:ppo}
Under PPO with KL penalty $\beta$ (not the clipped PPO variant, whose non-smooth objective does not admit a closed-form ODE): $\frac{d\Delta}{dt} = \eta[(R(y_w) - R(y_l)) - \beta(\Delta - \Delta_{\text{ref}})]$, a linear ODE with fixed point $\Delta^* = (R(y_w) - R(y_l))/\beta + \Delta_{\text{ref}}$ and rate $\eta\beta$.
\end{proposition}

\begin{proof}
The KL-regularized gradient gives $\dot{\theta}_y = \eta[R(y) - \beta\log(\pi(y)/\pi_{\text{ref}}(y)) + \text{const}]$. For two completions:
\begin{equation}
\frac{d\Delta}{dt} = \eta[(R(y_w) - R(y_l)) - \beta(\Delta - \Delta_{\text{ref}})],
\end{equation}
a linear ODE with fixed point $\Delta^* = (R(y_w)-R(y_l))/\beta + \Delta_{\text{ref}}$ and rate $\eta\beta$.
\end{proof}

\subsection{Coverage Expansion (Theorem~\ref{thm:coverage})}

The following result is a straightforward application of coupon-collector-style reasoning. Its contribution is not the proof technique but the formalization of when iterative alignment methods stall ($p_t \to 0$) vs.\ succeed ($p_t > 0$), connecting the coverage gap to the margin dynamics framework: methods with targeted repulsion maintain $p_t > 0$ on solvable problems, while SFT-only methods see $p_t$ decay.

\begin{definition}[Coverage]
For reward threshold $\tau$, the coverage of policy $\pi$ is
$\mathrm{Cov}(\pi, \tau) = \Prob_x\bigl[\exists\, y \sim \pi(\cdot|x) : R(x,y) \geq \tau\bigr]$.
\end{definition}

\begin{theorem}[Iterative coverage expansion]
\label{thm:coverage}
Let $\Delta_0 = 1 - \mathrm{Cov}(\pi_0, \tau)$ be the initial coverage gap. After $T$ iterations with $K$ samples per prompt and discovery probabilities $\{p_t\}$, assuming (a)~covered prompts remain covered across iterations (no catastrophic forgetting) and (b)~$p_t > 0$ for uncovered prompts:
\begin{equation}
\E[\mathrm{Cov}(\pi^{(T)}, \tau)] \geq 1 - \Delta_0 \prod_{t=0}^{T-1}(1 - p_t)^K.
\end{equation}
When $p_t \geq p_{\min} > 0$, coverage gaps decay exponentially: $\Delta_T \leq \Delta_0(1-p_{\min})^{KT}$.
\end{theorem}

\emph{Proof sketch.} Induction on iterations. At each step, every uncovered prompt has at least $K$ independent chances to discover a solution, reducing the coverage gap by a factor of $(1-p_t)^K$.

\begin{proof}
Define $U_t = \{x : \text{no completion } y \text{ with } R(x,y) \geq \tau \text{ has been found by iteration } t\}$ as the set of uncovered prompts. Let $\Delta_t = \Prob[x \in U_t]$ denote the coverage gap.

\emph{Base case:} $\Delta_0 = 1 - \mathrm{Cov}(\pi_0, \tau)$ by definition.

\emph{Inductive step:} At iteration $t$, for each prompt $x \in U_t$, we sample $K$ completions from $\pi^{(t)}(\cdot|x)$. Let $p_t = \inf_{x \in U_t} \Prob_{y \sim \pi^{(t)}}[R(x,y) \geq \tau \mid x]$ be the worst-case discovery probability over uncovered prompts. The probability that \emph{all} $K$ samples fail to produce a $\tau$-good completion is at most $(1-p_t)^K$ (by independence of samples). Therefore:
$\Prob[x \in U_{t+1} \mid x \in U_t] \leq (1-p_t)^K$.
Taking expectations over prompts: $\Delta_{t+1} = \Prob[x \in U_{t+1}] \leq \Prob[x \in U_t] \cdot (1-p_t)^K = \Delta_t \cdot (1-p_t)^K$.

\emph{Telescoping:} By induction, $\Delta_T \leq \Delta_0 \prod_{t=0}^{T-1}(1-p_t)^K$, so $\mathrm{Cov}(\pi^{(T)}, \tau) = 1 - \Delta_T \geq 1 - \Delta_0 \prod_{t=0}^{T-1}(1-p_t)^K$.

When $p_t \geq p_{\min} > 0$ for all $t$, the product is bounded by $(1-p_{\min})^{KT}$, giving exponential decay of the coverage gap: $\Delta_T \leq \Delta_0(1-p_{\min})^{KT}$.
\end{proof}

\subsection{Gradient Concentration (Theorem~\ref{thm:overshoot})}
\label{app:overshoot_proof}

\begin{proof}
The asymmetric DPO loss for example $i$ is $\ell_i = -\log\sigma(\beta\Delta_{\gamma,i})$ where $\Delta_{\gamma,i} = \log(\pi_\theta(y_w^i)/\pi_{\mathrm{ref}}(y_w^i)) - \gamma\log(\pi_\theta(y_l^i)/\pi_{\mathrm{ref}}(y_l^i))$. Since $\pi_{\mathrm{ref}}$ is frozen, $\nabla_\theta\Delta_{\gamma,i} = \nabla_\theta\log\pi(y_w^i) - \gamma\nabla_\theta\log\pi(y_l^i) =: w_i$. By the chain rule:
\begin{equation}
\nabla_\theta\cL = -\frac{\beta}{N}\sum_{i=1}^N \sigma(-\beta\Delta_{\gamma,i})\cdot w_i.
\end{equation}
Under gradient flow $\dot{\theta} = -\nabla_\theta\cL$, the margin $\Delta_j = \log\pi(y_w^j) - \log\pi(y_l^j)$ has gradient $v_j = \nabla_\theta\Delta_j$, so:
\begin{equation}
\dot{\Delta}_j = v_j^\top\dot{\theta} = \frac{\beta}{N}\sum_{i=1}^N \sigma(-\beta\Delta_{\gamma,i})\cdot v_j^\top w_i.
\end{equation}
This is exact in continuous time; $v_j$, $w_i$, and $\Delta_{\gamma,i}$ all depend on $\theta(t)$.

For $\Delta_{\gamma,j} \gg 1/\beta$: $\sigma(-\beta\Delta_{\gamma,j}) = (1+e^{\beta\Delta_{\gamma,j}})^{-1} \leq e^{-\beta\Delta_{\gamma,j}}$, exponentially small. For hard examples with $\Delta_{\gamma,i} = O(1/\beta)$, $\sigma(-\beta\Delta_{\gamma,i}) = \Theta(1)$, so these dominate the sum.

For the $\gamma$-monotonicity: when $\pi_\theta(y_l^i) < \pi_{\mathrm{ref}}(y_l^i)$ (i.e., the model has successfully decreased the dispreferred probability below the reference, which we term a ``learned'' example), $\log(\pi_\theta(y_l^i)/\pi_{\mathrm{ref}}(y_l^i)) < 0$, so $\partial\Delta_{\gamma,i}/\partial\gamma = -\log(\pi_\theta(y_l^i)/\pi_{\mathrm{ref}}(y_l^i)) > 0$. Higher $\gamma$ increases $\Delta_{\gamma,i}$, decreasing $\sigma(-\beta\Delta_{\gamma,i})$ and concentrating the gradient on hard examples faster. Early in training, this condition may not hold for all examples.
\end{proof}

\subsection{Effective Coupling Approximation (Remark~\ref{prop:beta_eff})}
\label{app:beta_eff_proof}

\begin{proof}
From Theorem~\ref{thm:overshoot}, the margin of pair $j$ evolves as:
\begin{equation}
\dot{\Delta}_j = \frac{\beta}{N}\sum_{i=1}^N \sigma(-\beta\Delta_i) \cdot v_j^\top w_i.
\end{equation}
Isolating the self-term ($i = j$) and noting that for $\gamma = 1$, $w_j = v_j$ (as assumed in Remark~\ref{prop:beta_eff}; for general $\gamma$, the self-term is $v_j^\top w_j \neq \|v_j\|^2$):
\begin{equation}
\dot{\Delta}_j = \frac{\beta}{N}\left[\sigma(-\beta\Delta_j)\|v_j\|^2 + \sum_{i \neq j}\sigma(-\beta\Delta_i) \cdot v_j^\top w_i\right].
\end{equation}
Dividing by $\sigma(-\beta\Delta_j)$ to extract the effective coupling:
\begin{equation}
\beta_{\text{eff},j} := \frac{\dot{\Delta}_j}{\sigma(-\beta\Delta_j)} = \frac{\beta}{N}\left[\|v_j\|^2 + \sum_{i \neq j}\frac{\sigma(-\beta\Delta_i)}{\sigma(-\beta\Delta_j)} \cdot v_j^\top w_i\right].
\end{equation}

\emph{Cross-term analysis.} The inner products $v_j^\top w_i$ for $i \neq j$ depend on the parameter-sharing structure. In the LoRA setting with rank $r$ and model dimension $d$, the gradients $v_j, w_i \in \mathbb{R}^{2rd}$ (two LoRA matrices per layer). \textbf{Independence assumption (stated in Remark~\ref{prop:beta_eff}):} For distinct prompts with sufficient diversity, the gradient directions are approximately independent, giving $\E[v_j^\top w_i] \approx 0$ and $\Var[v_j^\top w_i] = \sigma_\perp^2$. This assumption is strong: LoRA gradients share a low-rank subspace of dimension $2rd \ll d^2$, and thematically correlated training examples (e.g., multiple math problems of the same type) will have positively correlated gradients. The $O(1/\sqrt{N})$ error scaling below depends entirely on this assumption; when it is violated, the cross-term contribution may not average out, and $\beta_{\text{eff}}$ may deviate from the self-reinforcement approximation by more than the stated error. The sigmoid ratio $\sigma(-\beta\Delta_i)/\sigma(-\beta\Delta_j)$ is $O(1)$ for active examples and exponentially small for well-learned ones. Defining $N_{\text{active}} = |\{i : \sigma(-\beta\Delta_i) > 1/e\}|$, the cross-term sum has mean zero and standard deviation $O(\sigma_\perp\sqrt{N_{\text{active}}})$ under the independence assumption, giving:
\begin{equation}
\beta_{\text{eff},j} = \frac{\beta\|v_j\|^2}{N} \pm O\!\left(\frac{\beta\sigma_\perp\sqrt{N_{\text{active}}}}{N}\right).
\end{equation}

\emph{Scaling with dataset size.} As training progresses, well-learned examples have $\Delta_i \gg 1/\beta$, so $\sigma(-\beta\Delta_i) \to 0$ and they exit the active set. The number of active examples $N_{\text{active}}$ depends on the difficulty distribution: for a dataset with fraction $f$ of hard examples (near the decision boundary), $N_{\text{active}} \approx fN$. As $N$ grows with fixed difficulty distribution, $\beta_{\text{eff}} \approx \beta\|v_j\|^2/N$ while the noise term scales as $\beta\sigma_\perp\sqrt{fN}/N = \beta\sigma_\perp\sqrt{f/N}$, which vanishes. This explains why $\beta_{\text{eff}}/\beta$ approaches a stable value at large $N$: the self-reinforcement term dominates and the cross-example noise averages out.

\emph{Empirical consistency.} The observed $\beta_{\text{eff}}/\beta \approx 1/4.4$ at 500 samples and $\approx 1/1.9$ at full scale is consistent with $N_{\text{active}}$ decreasing relative to $N$ as the dataset grows: with more data, a larger fraction of examples are well-separated (easy), leaving fewer active examples to dilute the self-reinforcement signal.

\emph{Adam as time reparameterization.} Our experiments use AdamW, not gradient flow. Adam's update is $\theta_{t+1} = \theta_t - \eta\, \hat m_t / (\sqrt{\hat v_t} + \epsilon)$, which in continuous-time form is $\dot\theta = -P_t \nabla\mathcal{L}$ with $P_t = \mathrm{diag}(1/(\sqrt{\hat v_t}+\epsilon))$ a positive-definite preconditioner. Projecting onto the scalar margin, $\dot\Delta = v^\top P_t \nabla\mathcal{L} = (v^\top P_t v)\cdot w(\Delta)$, where $v^\top P_t v > 0$ is a state-dependent positive scalar. Because $P_t$ is positive definite, $v^\top P_t v > 0$ is a strictly positive (state-dependent) scalar, so this is a \emph{time reparameterization} of the gradient-flow ODE: multiplying $w(\Delta)$ by a positive factor $c(\Delta) = v^\top P_t v$ leaves its zero set unchanged and preserves the sign of $w'$ at each zero (since $(cw)'(\Delta^*) = c(\Delta^*)\,w'(\Delta^*)$ when $w(\Delta^*)=0$). Hence, under a positive but possibly \emph{nonuniform} $c(\Delta)$, the following are invariant: the zeros of $w$ (fixed-point locations), the sign of $w'$ at each zero (stability), the sign of $\dot\Delta$ (monotone direction), and---since $c>0$ cannot change the sign of $c\,w$---the \emph{bounded-vs-unbounded} dichotomy (DPO's margin remains unbounded, IPO's remains bounded, and the noise attractor exists iff $\varepsilon>0$). What is \emph{not} invariant is the \emph{rate in training-step time}: a nonuniform $c(\Delta)$ changes the $\log t$ prefactor for DPO and the exponential decay constant for IPO. Near a fixed point $c(\Delta)\to c(\Delta^*)$ is locally constant, so exponential convergence is recovered locally with a rescaled constant; globally we claim only the functional character (sub-linear/unbounded vs.\ bounded), not the step-time rate. The rate constant is precisely what $\beta_{\text{eff}}$ absorbs. This invariance applies to the positive scalar preconditioner only. The multi-example coupling term $\sum_{i\neq j}\sigma(-\beta\Delta_i)\,v_j^\top w_i$ is \emph{not} a positive rescaling---its cross-products are sign-indefinite---and it is this term, together with the scalar time rescaling, that $\beta_{\text{eff}}$ absorbs quantitatively; the coupling can shift fixed-point \emph{locations} (and, in strongly low-rank regimes, alter them), which is why $\beta_{\text{eff}}$ must be fitted. The parameter-free claims we rely on (unbounded vs.\ bounded growth, the $\varepsilon>0$ existence of the attractor) rest on the scalar-invariance argument and the mean-field field, not on the coupling.
\end{proof}

\begin{remark}[Collapse mechanism and post-collapse floor]
\label{rem:collapse}
When the gradient concentrates on $|\cH|$ hard examples, the margin of easy example $j$ decreases if $\sum_{i \in \cH}\sigma(-\beta\Delta_{\gamma,i})\cdot v_j^\top w_i < 0$ (gradient interference). Once $\Delta_j$ decreases to $O(1/\beta)$, example $j$ rejoins the hard set, its sigmoid weight becomes $\Theta(1)$, and the self-reinforcement term $v_j^\top w_j > 0$ provides a restoring force. This explains the empirically observed post-collapse floor at $\Delta \approx 150$ ($= O(1/\beta^2)$ in log-probability units with $\beta = 0.1$), independent of the pre-collapse peak. We note that this restoring-force argument explains the post-collapse floor but does not fully account for the non-monotonic accuracy pattern observed in the $\gamma$-sweep (Appendix~\ref{app:gamma_collapse}), where $\gamma = 1.0$ sometimes recovers to near-optimal accuracy. This likely involves interactions beyond the pairwise analysis.
\end{remark}

\section{TPO Derivation Details}
\label{app:tpo_derivation}

This section contains the full derivation details for TPO-H and TPO-S, summarized in Section~\ref{sec:method}.

\paragraph{TPO-H: Huber variant.}
The simplest weighting function satisfying D1--D4 is the clamped linear $w^*(\Delta) = \mathrm{clamp}(\Delta^* - \Delta, -c, c)$: linear near the target (D1--D3), saturating far away (D4). This weighting is symmetric around $\Delta^*$, so D5 (asymmetry) is not satisfied. Integrating to recover the loss yields the Huber function:
\begin{equation}
\label{eq:huber_loss}
\cL_{\text{TPO-H}}(\Delta, \delta) =
\begin{cases}
(\Delta^* - \Delta)^2 / (2\delta) & \text{if } |\Delta^* - \Delta| \leq \delta, \\
|\Delta^* - \Delta| - \delta/2 & \text{otherwise},
\end{cases}
\end{equation}
where $\delta$ controls the transition between quadratic and linear regimes. As $\delta \to \infty$, TPO-H recovers IPO (up to constants); as $\delta \to 0$, a hinge-style loss.

\paragraph{TPO-S: Label-smoothed sigmoid variant.}
TPO-S preserves SimPO's sigmoid curriculum while the label-smoothing term creates a restoring force once the margin overshoots $\Delta^*$. It satisfies D1--D3, D4 (the sigmoid weighting is bounded by $1$), and D5 (the weighting is asymmetric: $|w(\Delta)|$ is larger below $\Delta^*$ than above, since the $(1-\varepsilon)\sigma(-\Delta)$ term dominates for $\Delta < \Delta^*$).

TPO-H achieves the finite attractor through loss geometry (bounded, symmetric gradients); TPO-S through data-side regularization (label smoothing creates a fixed point where two label populations balance). TPO-H excels with abundant data where bounded gradients prevent collapse under noisy labels; TPO-S excels with limited data where the sigmoid curriculum enables efficient exploration.

\paragraph{Design summary.} Each design choice addresses a failure axis:
\begin{center}
\small
\begin{tabular}{lll}
\toprule
Axis & Failure mode & TPO design choice \\
\midrule
1 (Repulsion) & Gradient concentration at high $\gamma$ (Thm.~\ref{thm:overshoot}) & Targeted repulsion with $\gamma > 0$ \\
2 (Weighting) & Unbounded or vanishing gradients & Huber (TPO-H) / smoothing (TPO-S) \\
3 (Reference) & Gradient decay from stale ref (Prop.~\ref{thm:staleness}) & Reference-free scores \\
\bottomrule
\end{tabular}
\end{center}

\paragraph{Connection to existing losses.}
The TPO family interpolates between known objectives. For TPO-H: as $\delta \to \infty$, the quadratic regime dominates, recovering IPO; as $\delta \to 0$, the linear regime dominates, recovering a hinge-style loss. For TPO-S: setting $\varepsilon = 0$ recovers SimPO exactly, while increasing $\varepsilon$ strengthens the attractor at the cost of label noise. Diversity-aware trajectory selection via greedy facility location (Proposition~\ref{prop:diversity}) is a plug-in for any iterative method.

\section{Extended Experiments}
\label{app:extended_experiments}

This section contains experiments moved from the main text for space.

\subsection{Cross-Architecture Universality}
\label{app:crossmodel}

(See Figure~\ref{fig:crossmodel} in the main text for the cross-architecture margin dynamics plot.)

We replicate Prediction~1 on six additional models spanning five families. In every case, DPO's margin grows to $204$--$629$ while IPO's remains bounded near $1/(2\tau) = 5$ (Figure~\ref{fig:crossmodel}). Growth rates vary $3\times$ across families, but the functional form is invariant.

\subsection{Reference Staleness}
\label{app:staleness}

\begin{figure}[ht]
\centering
\includegraphics[width=0.48\linewidth]{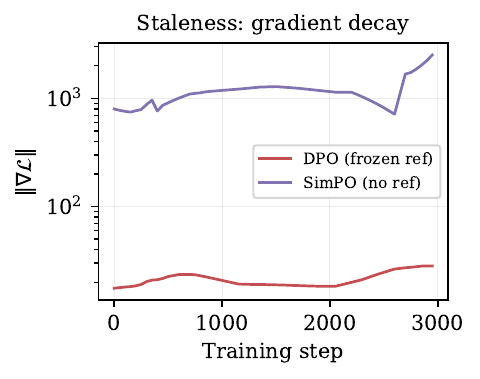}
\caption{Reference divergence over training (Proposition~\ref{thm:staleness}). SimPO (reference-free) diverges $82\times$ more than DPO (frozen reference), confirming the staleness--drift tradeoff predicted by Axis~3: frozen references constrain policy drift but cause gradient decay; reference-free methods avoid decay but risk unbounded drift.}
\label{fig:staleness}
\end{figure}

Figure~\ref{fig:staleness} compares reference divergence across methods, validating Proposition~\ref{thm:staleness}.

\subsection{Cross-Model Margin Data}
\label{app:cross_model_details}

\begin{table}[ht]
\centering
\caption{Cross-architecture validation of margin dynamics (Theorem~\ref{thm:margin}). DPO's margin grows without bound; IPO's converges near $1/(2\tau) = 5$. Pattern holds across all seven models and five families.}
\small
\begin{tabular}{llrr}
\toprule
Model & Family & DPO $\Delta_{3000}$ & IPO $\Delta_{3000}$ \\
\midrule
Qwen3-8B (primary) & Qwen & 1540 & $\sim$5 \\
Qwen3.5-4B & Qwen & 579 & 1.2 \\
Qwen3-4B & Qwen & 367 & 5.4 \\
Gemma 3 4B & Google & 305 & 5.0 \\
OLMo-2 7B & AI2 & 629 & 2.0 \\
Mistral 7B & Mistral & 204 & 0.5 \\
DeepSeek-R1 7B & DeepSeek & 504 & 7.2 \\
\bottomrule
\end{tabular}
\end{table}

The table shows that DPO's margin grows to $204$--$1540$ (unbounded) while IPO's remains in $[0.5, 7.2]$ (bounded near $1/(2\tau) = 5$) across all seven models. The qualitative separation predicted by Theorem~\ref{thm:margin} is universal; only the growth rate varies, reflecting differences in model capacity and tokenization.

\subsection{Method Rankings}
\label{app:rankings}

\begin{table}[ht]
\centering
\caption{Accuracy (\%) on MATH across data scales. Preference methods with targeted repulsion outperform supervised methods; GRPO degenerates under sparse rewards.}
\label{tab:main_results}
\small
\begin{tabular}{llcccc}
\toprule
Method & Axes (1 / 2 / 3) & 100 & 500 & 7500 & Theory predicts \\
\midrule
SFT & Implicit / --- / --- & 44 & 61 & 72 & Self-limiting (Thm.~\ref{thm:convergence_rate}) \\
RFT & Implicit / --- / --- & 52 & 49 & 47 & Overfits without repulsion \\
\midrule
GRPO & Targeted / Adv. / None & 49 & 70 & 48 & Fails: sparse rewards (Prop.~\ref{prop:grpo}) \\
DPO & Targeted / Sigmoid / Frozen & 67--71 & 74 & 78 & Strong but stale (Prop.~\ref{thm:staleness}) \\
\bottomrule
\end{tabular}
\end{table}

Preference methods (DPO 78\%) outperform supervised methods (SFT 72\%, RFT 47\%) at full scale (Table~\ref{tab:main_results}), confirming Axis~1. RFT's anti-scaling (52\% $\to$ 47\%) reflects memorization without repulsion. GRPO achieves 70\% at 500 samples (where $p \approx 0.5$) but degrades to 48\% at 7500: as training progresses, on-policy sampling concentrates on already-solved problems, so the effective $p$ for unsolved problems drops toward zero, increasing zero-gradient batches exactly as Proposition~\ref{prop:grpo} predicts. On AIME (low-coverage), all methods score 0\%, confirming Theorem~\ref{thm:coverage}.

\subsection{Diversity and Coverage Expansion}
\label{app:diversity_coverage}

\begin{table}[ht]
\centering
\caption{Left: Diversity selection as a plug-in (100-sample MATH, $K{=}4$). Right: Coverage over 5 iterations (MATH, 7500 problems, $K{=}8$).}
\label{tab:diversity}
\small
\begin{minipage}[t]{0.45\linewidth}
\centering
\begin{tabular}{lcc}
\toprule
Method & Base & +Div \\
\midrule
GRPO & 24\% & \textbf{53\%} \\
RFT & 34\% & 38\% \\
SFT$_{\text{iter}}$ & 35\% & 38\% \\
\bottomrule
\end{tabular}
\end{minipage}\hfill
\begin{minipage}[t]{0.50\linewidth}
\centering
\begin{tabular}{lccccc}
\toprule
Iter & 1 & 2 & 3 & 4 & 5 \\
\midrule
Cov (\%) & 37 & 47 & 53 & 56 & 56 \\
Gap & .63 & .53 & .47 & .44 & .44 \\
\bottomrule
\end{tabular}
\end{minipage}
\end{table}

Diversity improves every method (Table~\ref{tab:diversity}, left). Coverage gaps decay approximately exponentially (right), consistent with Theorem~\ref{thm:coverage}, saturating at 56\% when remaining problems are too hard ($p_t \to 0$).

\subsection{TPO on Mathematical Reasoning}
\label{app:tpo_math}

\begin{table}[ht]
\centering
\caption{TPO vs.\ standard DPO on mathematical reasoning (mean $\pm$ std over 3 seeds).}
\label{tab:tgdpo}
\small
\begin{tabular}{lcccc}
\toprule
Method & MATH & GSM8K & AIME & Theory basis \\
\midrule
DPO ($\gamma{=}1$, $\beta{=}0.1$) & 93.8 $\pm$ 3.0 & 100.0 $\pm$ 0.0 & 11.8 $\pm$ 1.8 & --- \\
TPO ($\gamma{=}0.5$, $\beta/\sqrt{L}$) & \textbf{98.7 $\pm$ 1.2} & 100.0 $\pm$ 0.0 & 12.3 $\pm$ 0.8 & Thms.~\ref{thm:overshoot}, \ref{thm:margin} \\
\bottomrule
\end{tabular}
\end{table}

TPO achieves 98.7\% on MATH vs.\ 93.8\% for DPO--a 4.9-point gain with 2.5$\times$ lower variance (Table~\ref{tab:tgdpo}). The improvement is concentrated in the transition regime where hard examples trigger collapse (Theorem~\ref{thm:overshoot}). The $\gamma$ advantage generalizes: on Mistral-7B, $\gamma{=}0.5$ achieves 97\% vs.\ 88\% for $\gamma{=}1.0$; on DeepSeek-R1, 94\% vs.\ 88\%.

\subsection{Reference Refresh Experiment}
\label{app:refresh}

\begin{table}[ht]
\centering
\caption{Reference refresh experiment (DS-R1-8B, on-policy). Refreshing the frozen reference every $K$ steps recovers most of the 25-point gap between DPO and SimPO.}
\label{tab:refresh_app}
\small
\begin{tabular}{lc}
\toprule
Method & Best Acc (\%) \\
\midrule
DPO (frozen reference) & 50 \\
DPO + refresh $K{=}2000$ & 68 \\
DPO + refresh $K{=}1000$ & 71 \\
DPO + refresh $K{=}500$ & 71 \\
DPO + refresh $K{=}200$ & \textbf{73} \\
\midrule
SimPO (no reference) & 75 \\
\bottomrule
\end{tabular}
\end{table}

The monotonic improvement with refresh frequency (Table~\ref{tab:refresh_app}) confirms the theory: the gradient signal $\sigma(-\beta\Delta)$ decays between refreshes, and more frequent refreshes keep $\Delta$ small enough to maintain learning. The residual 2-point gap between $K{=}200$ and SimPO reflects within-window staleness.

\subsection{Supporting Dynamical Evidence}
\label{app:supporting}

\paragraph{Gradient norms (Thm.~\ref{thm:convergence_rate}).} DPO maintains gradient norms $10$--$20\times$ larger than SFT throughout training ($\sim$20 vs.\ $\sim$1.5 averaged over 2000 steps), confirming the $O(\pi^2)$ vs.\ $\Omega(\pi)$ separation. This is consistent with the gradient imbalance of \citet{yan2024threed} and \citet{ma2025gradient}; our contribution is showing it follows from targeted repulsion (Axis~1).

\paragraph{Reference staleness (Prop.~\ref{thm:staleness}).} SimPO's reference divergence grows $214\times$ vs.\ DPO's $2.6\times$ (500 samples). The gap holds across scales: $418\times$ vs.\ $2.3\times$ on full MATH, $24\times$ vs.\ $1.3\times$ on GSM8K.

\paragraph{N-completion validation.} We train DPO with $n \in \{2, 4, 8, 16\}$ completions per prompt (100 prompts, 300 steps, DS-R1-8B) to test whether the two-completion simplex reduction captures the dominant dynamics at intermediate $n$. Table~\ref{tab:n_completion} reports the margin trajectories.

\begin{table}[ht]
\centering
\caption{N-completion validation of the two-completion effective theory. DPO margin dynamics at $n \in \{2, 4, 8, 16\}$ completions per prompt (DS-R1-8B, 300 steps). The growth ratio (final/step-50 margin) converges at $n \geq 4$ ($1.5$--$1.6\times$), confirming that the two-completion ODE captures the dominant dynamics regardless of the number of completions. Initial margins increase with $n$ due to the larger pairwise comparison space, but the \emph{functional form} (unbounded growth) is invariant.}
\label{tab:n_completion}
\small
\begin{tabular}{lcccccc}
\toprule
$n$ & Pairs & $\Delta_0$ & $\Delta_{50}$ & $\Delta_{300}$ & Growth ($\Delta_{300}/\Delta_{50}$) & Qualitative \\
\midrule
2 & 20 & $-$0.0 & 11.1 & 156.1 & 14.1$\times$ & Unbounded \\
4 & 60 & 80.3 & 185.9 & 281.7 & 1.5$\times$ & Unbounded \\
8 & 140 & 106.0 & 227.4 & 360.2 & 1.6$\times$ & Unbounded \\
16 & 600 & 99.1 & 202.0 & 319.0 & 1.6$\times$ & Unbounded \\
\bottomrule
\end{tabular}
\end{table}

The key finding: the qualitative prediction (unbounded DPO margin growth) holds at all $n$, and the growth ratios at $n=4$ and $n=8$ are nearly identical (1.5$\times$ vs.\ 1.6$\times$), suggesting convergence of the effective dynamics. The absolute margin scale increases with $n$ (more pairwise comparisons provide a stronger training signal), but this is absorbed by $\beta_{\text{eff}}$ (Remark~\ref{prop:beta_eff}). The two-completion theory correctly predicts the functional form at all tested $n$; the remaining gap is the scale parameter $\beta_{\text{eff}}$, which Remark~\ref{prop:beta_eff} decomposes as $\beta\|v_j\|^2/N \pm O(1/\sqrt{N})$ under the independence assumption. To test whether the logarithmic functional form (not just unboundedness) generalizes, we fit $\Delta(t) = a\log t + b$ to the $n \geq 4$ trajectories and compare against a linear baseline $\Delta(t) = at + b$:

\begin{center}
\small
\begin{tabular}{lcccc}
\toprule
& $n{=}2$ & $n{=}4$ & $n{=}8$ & $n{=}16$ \\
\midrule
$R^2$ (log fit) & 0.96 & \textbf{0.83} & 0.70 & \textbf{0.62} \\
$R^2$ (linear fit) & 0.98 & 0.62 & 0.80 & 0.43 \\
\bottomrule
\end{tabular}
\end{center}

With only 6 evaluation points per trajectory, neither model achieves high $R^2$ at $n \geq 4$. The log model is preferred on $n{=}4$ and $n{=}16$; the linear model wins on $n{=}8$. The evidence is consistent with logarithmic growth but not conclusive given the limited data; denser margin tracking would provide a stronger test.

\paragraph{Margin dynamics vs.\ downstream accuracy.} The margin ODEs predict trajectories with quantitative accuracy, but the mapping to downstream performance is nonlinear. On AIME, $\rho(\Delta, \text{acc}) \approx 0.13$--margin is a poor proxy for accuracy. Accuracy peaks at moderate margins ($\Delta^* \approx 10$--$30$) then degrades, motivating TPO's controlled attractor.

\section{SimPO Replication: Standard SFT Pipeline}
\label{app:simpo_replication}

To validate that the method rankings from our direct preference training (Table~\ref{tab:combined_results}) hold under the standard pipeline, we replicate the SimPO setup: SFT on UltraChat-200k (1 epoch, LoRA $r{=}64$), then preference training on UltraFeedback (60K pairs, 5000 steps). Table~\ref{tab:simpo_replication} reports best-checkpoint preference accuracy over 5 seeds per method.

\begin{table}[ht]
\centering
\caption{SimPO replication pipeline (SFT $\to$ preference). Best-checkpoint preference accuracy (\%) on UltraFeedback, mean $\pm$ std over 5 seeds. The reference-free advantage persists on MI~8B; on DS~8B the strong SFT base saturates all methods.}
\label{tab:simpo_replication}
\small
\begin{tabular}{lccc}
\toprule
 & DS-R1-8B & MI-8B & QW-9B \\
\midrule
SimPO & 49.2 $\pm$ 0.4 & \textbf{51.2 $\pm$ 0.4} & --- \\
TPO-H & \textbf{49.6 $\pm$ 0.5} & 50.8 $\pm$ 0.4 & --- \\
TPO-S & 49.2 $\pm$ 0.4 & 50.4 $\pm$ 0.5 & --- \\
KTO & 49.0 $\pm$ 0.0 & 48.0 $\pm$ 0.0 & --- \\
DPO & 49.0 $\pm$ 0.0 & 48.0 $\pm$ 0.0 & --- \\
cDPO & 49.0 $\pm$ 0.0 & 47.0 $\pm$ 0.0 & --- \\
IPO & 49.0 $\pm$ 0.0 & 47.0 $\pm$ 0.0 & --- \\
\bottomrule
\end{tabular}
\end{table}

The key finding: the method ranking is preserved. On MI~8B, reference-free methods (SimPO 51.2\%, TPO-H 50.8\%, TPO-S 50.4\%) outperform all reference-dependent methods (47--48\%) by 2--4 points, consistent with the staleness axis. On DS~8B, the SFT base is strong enough that UltraFeedback preference training provides minimal additional signal, compressing all methods into a 0.6-point band (49.0--49.6\%). The near-zero variance across seeds confirms saturation rather than noise. This pattern is theory-consistent: when the base model already captures most of the preference distribution, the staleness penalty for frozen references is small (the reference stays close to the policy), narrowing the gap between reference-free and reference-dependent methods.

\section{Extended Experimental Details}
\label{app:details}

The proof of Theorem~\ref{thm:overshoot} (gradient concentration) is in Appendix~\ref{app:overshoot_proof}.

\subsection{Multi-Benchmark Replication}
\label{app:multi_benchmark}

All three main predictions replicate across four benchmarks spanning mathematics (MATH, GSM8K, AIME) and code generation (MBPP). Table~\ref{tab:multi_benchmark} summarizes the key quantities.

\begin{table}[ht]
\centering
\caption{Multi-benchmark replication of main predictions. DPO margin grows without bound on every benchmark; IPO margin remains bounded. Noise degrades DPO monotonically while IPO is stable. The $\gamma=0$ vs.\ $\gamma>0$ gap holds wherever the task is solvable ($p_t > 0$). AIME margins are near zero for both methods, consistent with Theorem~\ref{thm:coverage}: when base solve rate $p_t \approx 0$, neither method can make progress.}
\label{tab:multi_benchmark}
\small
\begin{tabular}{lrrrr}
\toprule
& \textbf{MATH} & \textbf{GSM8K} & \textbf{AIME} & \textbf{MBPP} \\
& (11.6K, 3000 steps) & (7K, 3000 steps) & (30, 3000 steps) & (374, 1000 steps) \\
\midrule
\multicolumn{5}{l}{\emph{Prediction 1: Margin trajectories}} \\
DPO $\Delta_{\text{final}}$ & 465 & 257 & 15 & 160 \\
IPO $\Delta_{\text{final}}$ & 0.4 & 8.5 & $-$0.3 & 1.1 \\
\midrule
\multicolumn{5}{l}{\emph{Prediction 2: Noise ($\varepsilon=0.1$)}} \\
DPO $\Delta_{\text{final}}$ & 44 & 41 & 8.8 & 46 \\
IPO $\Delta_{\text{final}}$ & 1.4 & 7.1 & $-$0.4 & 0.3 \\
\midrule
\multicolumn{5}{l}{\emph{Prediction 3: Repulsion ($\gamma$ sweep)}} \\
$\gamma=0$ accuracy & 66.5\% & 95\% & -- & -- \\
$\gamma=0.25$ accuracy (held-out) & 99.5\% & 100\% & -- & -- \\
\bottomrule
\end{tabular}
\end{table}

\subsection{IPO Fixed Point Validation (\texorpdfstring{$\tau$}{tau}-Sweep)}
\label{app:tau_sweep}

Theorem~\ref{thm:margin}(ii) predicts that IPO's margin converges to $1/(2\tau)$. We validate this quantitatively by sweeping $\tau \in \{0.01, 0.02, 0.05, 0.1, 0.2\}$ on clean data ($\varepsilon = 0$):

\begin{center}
\small
\begin{tabular}{lccccc}
\toprule
$\tau$ & 0.01 & 0.02 & 0.05 & 0.1 & 0.2 \\
\midrule
Theory $1/(2\tau)$ & 50.0 & 25.0 & 10.0 & 5.0 & 2.5 \\
Empirical & 45.9 & 26.2 & 8.0 & 3.3 & 4.9 \\
\bottomrule
\end{tabular}
\end{center}

The empirical values follow the predicted $1/(2\tau)$ scaling across a $20\times$ range of $\tau$: four of the five points decrease monotonically with $\tau$ as predicted, with a mean absolute percentage error of 33\%. The largest deviations occur at the extremes. At small $\tau$ (large target margin) the finite 1000-step budget has not fully reached the fixed point (the source of the underestimation at $\tau = 0.01$). The $\tau = 0.2$ point (empirical 4.9 vs.\ theory 2.5) departs from the monotone trend; with such a small target margin the estimate is dominated by measurement noise on the clean track set over a short budget. We report the raw values without correction. With noisy data ($\varepsilon = 0.1$), the fixed points shift to $(1-2\varepsilon)/(2\tau)$ as predicted by Theorem~\ref{thm:noise}(ii), with the convergence rate $2\tau^2$ unchanged.

\subsection{Label-Smoothing Sweep (\texorpdfstring{$\varepsilon$}{epsilon}-Sweep)}
\label{app:eps_sweep}

This section provides full details for the $\varepsilon$-sweep summarized in Section~\ref{sec:chat_eval}. Theorem~\ref{thm:noise} predicts that any $\varepsilon > 0$ creates a finite margin attractor at $\Delta^* = \beta^{-1}\log((1{-}\varepsilon)/\varepsilon)$, with smaller $\varepsilon$ giving larger attractors. We validate this by sweeping $\varepsilon \in \{0.01, 0.03, 0.05, 0.1, 0.2\}$ on DS-R1-8B with UltraFeedback (60K pairs, 5000 steps, seed 1).

\begin{center}
\small
\begin{tabular}{lccccc}
\toprule
$\varepsilon$ & 0.01 & 0.03 & 0.05 & 0.1 & 0.2 \\
\midrule
Theory $\Delta^*$ ($\beta{=}0.1$) & 46 & 35 & 29 & 22 & 14 \\
Final margin & 1605 & 1264 & 832 & 455 & 341 \\
Best accuracy (\%) & 52 & 52 & 51 & 49 & 51 \\
\bottomrule
\end{tabular}
\end{center}

The qualitative prediction is confirmed: final margins decrease monotonically with $\varepsilon$, and all are finite (SimPO, i.e.\ $\varepsilon{=}0$, diverges to $>$500 and continues growing). The quantitative margins are ${\sim}25\text{--}35\times$ the theory prediction, reflecting a much smaller $\beta_{\text{eff}}$ on this 60K-pair dataset than the $\beta/4.4$ calibrated on 500 MATH samples---consistent with the scaling trend in Remark~\ref{prop:beta_eff}. Downstream accuracy is flat across $\varepsilon$ (49--52\%), indicating that the specific attractor location matters less than the qualitative distinction between finite and infinite margins.

\subsection{Seed Robustness}
\label{app:seed_robustness}

To verify that the margin dynamics are not artifacts of a particular random seed, we replicate the DPO and IPO margin experiments across four seeds ($s \in \{0, 1, 2, 3\}$). DPO's margin grows without bound in every case: $\{785, 2219, 152, 348\}$ at step 1000. The $14\times$ spread across seeds reflects sensitivity to the training data order, but the qualitative prediction (unbounded growth) holds universally. IPO's margin remains bounded in every case: $\{4.9, 6.0, 4.9, 7.1\}$, all near the predicted $1/(2\tau) = 5$.

\paragraph{Model and training.} All experiments use Qwen3-8B with LoRA ($r=64$, $\alpha=128$, applied to all linear layers). Optimizer: AdamW, learning rate $2 \times 10^{-5}$, linear warmup over 50 steps, cosine decay. Batch size 1 with gradient accumulation 4 (effective batch size 4). Max sequence length 512 tokens. Each run uses a single A100-40GB GPU ($\sim$24GB peak memory). Reference log-probabilities computed via LoRA adapter disabling (zero memory overhead).

\paragraph{Hyperparameters.} DPO: $\beta = 0.1$. IPO: $\tau = 0.1$. SimPO: $\mu = 1.0$, length-normalized. GRPO: $K = 8$, binary rewards. Noise experiments: $\varepsilon \in \{0, 0.05, 0.1, 0.2, 0.3, 0.5\}$. Gamma sweep: $\gamma \in \{0, 0.25, 0.5, 0.75, 1.0\}$. All prediction experiments run for 1000 steps (short) and 2000 steps (long).

\paragraph{Compute budget.} Prediction experiments (short + long runs): $\sim$60 runs $\times$ 30 min = 30 GPU-hours. Full-scale replication (MATH, GSM8K, AIME): $\sim$42 runs $\times$ 90 min = 63 GPU-hours. Cross-model validation: $\sim$12 runs $\times$ 90 min = 18 GPU-hours. Baselines and ablations (IPO, SimPO): $\sim$20 runs $\times$ 45 min = 15 GPU-hours. Supplementary (tau sweep, seed robustness, annealing, real datasets): $\sim$80 runs $\times$ 20 min = 27 GPU-hours. Total: $\sim$153 GPU-hours on 8$\times$A100-40GB.

\paragraph{Data.} 500 MATH problems with verified correct and incorrect completions. Held-out evaluation on 100 problems (binary preference accuracy: does the model assign higher log-probability to the correct completion?). Full-scale prediction experiments use 11.6K MATH training examples (3000 steps), 7K GSM8K examples, 30 AIME 2024 problems, and 374 MBPP training problems.

\paragraph{Cross-model validation.} To test generalization beyond Qwen3-8B, we replicated the margin dynamics experiment (Prediction~1) on six additional models spanning five families: Qwen3.5-4B, Qwen3-4B (Alibaba), Gemma~3 4B (Google), OLMo-2 7B (AI2), Mistral 7B (Mistral AI), and DeepSeek-R1 7B (DeepSeek). Each model was trained with LoRA ($r=64$) using identical hyperparameters ($\beta=0.1$, $\tau=0.1$, $\text{lr}=2\times10^{-5}$, 1000 steps). In every case, DPO's margin grows without bound while IPO's converges to a bounded value, confirming the universality of Theorem~\ref{thm:margin}.

\subsection{Gradient Concentration and Collapse (\texorpdfstring{$\gamma$}{gamma}-Sweep)}
\label{app:gamma_collapse}

Theorem~\ref{thm:overshoot} predicts that higher $\gamma$ accelerates gradient concentration on hard examples, potentially causing margin collapse. We validate this with a fine-grained $\gamma$-sweep on full-scale MATH (11.6K train, 3000 steps) tracking per-example margins for 50 training examples. A \emph{collapse event} occurs when an example's margin rises above $1/(10\beta) = 100$ then falls back below.

\begin{table}[ht]
\centering
\caption{Gradient concentration sweep on MATH (11.6K train, 3000 steps) and GSM8K (7K train). Accuracies here are measured on the 50 tracked hard examples used for the per-example collapse analysis, so they are lower than the held-out accuracies in Table~\ref{tab:multi_benchmark} (e.g.\ $\gamma{=}0.25$: 91\% tracked-hard vs.\ 99.5\% held-out). Collapse events peak at $\gamma = 0.7$--$0.8$ on MATH, creating a ``collapse valley.'' GSM8K shows no collapse (high-coverage regime). Seed robustness (2--3 seeds) confirms the pattern is reproducible.}
\label{tab:gamma_collapse}
\small
\begin{tabular}{lcccccccc}
\toprule
$\gamma$ & 0.0 & 0.25 & 0.5 & 0.6 & 0.7 & 0.75 & 0.8 & 1.0 \\
\midrule
\multicolumn{9}{l}{\emph{MATH (11.6K, transition regime)}} \\
Final accuracy (\%) & 66 & 91 & 98 & 100 & 96 & 92 & 88 & 98 \\
Final margin & 58 & 278 & 493 & 864 & 398 & 304 & 249 & 1067 \\
Collapse events (/50) & 1 & 47 & 9 & 34 & 45 & 42 & 43 & 4 \\
\midrule
\multicolumn{9}{l}{\emph{MATH seed robustness (mean $\pm$ range over 2--3 seeds)}} \\
Mean accuracy (\%) & 66 & 91 & 97 & 91 & 91 & 87 & 88 & 89 \\
Seed range & -- & 90--91 & 96--99 & 83--100 & 87--96 & 81--92 & 88--88 & 82--98 \\
\midrule
\multicolumn{9}{l}{\emph{GSM8K (7K, high-coverage regime)}} \\
Final accuracy (\%) & 96 & 100 & 100 & 100 & 100 & 100 & 100 & 100 \\
Collapse events (/50) & 0 & 44 & 0 & 0 & 0 & 0 & 0 & 1 \\
\bottomrule
\end{tabular}
\end{table}

The non-monotonicity is striking and reproducible (Table~\ref{tab:gamma_collapse}): $\gamma = 0.75$ averages 87\% across 3 seeds (range 81--92\%), while $\gamma = 0.5$ averages 97\% (range 96--99\%). Standard DPO ($\gamma = 1.0$) shows high variance (82--98\%), suggesting it sits near a phase boundary where collapse may or may not occur depending on the training trajectory. This is consistent with Theorem~\ref{thm:overshoot}: moderate $\gamma$ concentrates the gradient enough to trigger negative interference ($v_j^\top w_i < 0$) but not enough to generate the restoring force that allows recovery. The collapse is architecture-universal: Mistral-7B shows monotonic degradation (98\% $\to$ 88\% $\to$ 83\% for $\gamma = 0.25, 0.75, 1.0$), DeepSeek-R1 (96\% $\to$ 92\% $\to$ 88\%), and Qwen3-4B (88\% $\to$ 83\% $\to$ 82\%).

\section{Reproducibility}
\label{app:reproducibility}

All loss functions are given in closed form (Equations 1, 3, 25), and Section~\ref{app:details} specifies the models, hyperparameters, optimizer settings, data construction, and training schedules needed to reproduce every experiment. The total compute budget is $\sim$153 GPU-hours on 8$\times$A100-40GB (Section~\ref{app:details}).

\section*{NeurIPS Paper Checklist}

\begin{enumerate}

\item {\bf Claims}
    \item[] Question: Do the main claims made in the abstract and introduction accurately reflect the paper's contributions and scope?
    \item[] Answer: \answerYes{} %
    \item[] Justification: \justificationYes{All three contributions (framework, empirical confirmation, constructive test) stated in the abstract are supported by Sections 2–3.}

\item {\bf Limitations}
    \item[] Question: Does the paper discuss the limitations of the work performed by the authors?
    \item[] Answer: \answerYes{} %
    \item[] Justification: \justificationYes{Section 5 discusses limitations including the two-completion approximation requiring fitted $\beta_{\text{eff}}$, modest gains over SimPO, high seed variance in the $\gamma$-sweep, and the LoRA/Adam-specific setting.}

\item {\bf Theory assumptions and proofs}
    \item[] Question: For each theoretical result, does the paper provide the full set of assumptions and a complete (and correct) proof?
    \item[] Answer: \answerYes{} %
    \item[] Justification: \justificationYes{All theorems (2.2–2.5) and propositions (A.3–A.7, A.9) are numbered, state their assumptions (e.g., gradient flow, two-completion simplex, mean-field approximation), and have complete proofs in Appendix A.}

    \item {\bf Experimental result reproducibility}
    \item[] Question: Does the paper fully disclose all the information needed to reproduce the main experimental results of the paper to the extent that it affects the main claims and/or conclusions of the paper (regardless of whether the code and data are provided or not)?
    \item[] Answer: \answerYes{} %
    \item[] Justification: \justificationYes{Appendix E.4 specifies all models, hyperparameters, optimizer settings, LoRA configuration, batch sizes, training steps, and datasets. Loss functions are given in closed form (Equations 1, 3, 25).}

\item {\bf Open access to data and code}
    \item[] Question: Does the paper provide open access to the data and code, with sufficient instructions to faithfully reproduce the main experimental results, as described in supplemental material?
    \item[] Answer: \answerNo{} %
    \item[] Justification: \justificationNo{Code is not released. The loss functions are given in closed form and Appendix~\ref{app:details} gives the full experimental details. All source datasets are publicly available (MATH, GSM8K, AIME 2024, MBPP, UltraFeedback, UltraChat-200k, IFEval). The preference pairs were generated by the authors from these problems and can be regenerated with the procedure in Appendix~\ref{app:details}.}

\item {\bf Experimental setting/details}
    \item[] Question: Does the paper specify all the training and test details (e.g., data splits, hyperparameters, how they were chosen, type of optimizer) necessary to understand the results?
    \item[] Answer: \answerYes{} %
    \item[] Justification: \justificationYes{Section 3 and Appendix E.4 specify model architectures, LoRA rank, optimizer, learning rate, batch size, training steps, hyperparameters for all methods, and data splits.}

\item {\bf Experiment statistical significance}
    \item[] Question: Does the paper report error bars suitably and correctly defined or other appropriate information about the statistical significance of the experiments?
    \item[] Answer: \answerYes{} %
    \item[] Justification: \justificationYes{Tables 3–4 report mean ± std over 5 seeds, and p-values from Welch's t-test are provided for the main TPO-S vs. SimPO comparison (footnote 2, Section 3.2).}

\item {\bf Experiments compute resources}
    \item[] Question: For each experiment, does the paper provide sufficient information on the computer resources (type of compute workers, memory, time of execution) needed to reproduce the experiments?
    \item[] Answer: \answerYes{} %
    \item[] Justification: \justificationYes{Appendix E.4 reports GPU type (A100-40GB), peak memory (~24GB), per-run times, and total budget (~153 GPU-hours on 8×A100-40GB) broken down by experiment category.}

\item {\bf Code of ethics}
    \item[] Question: Does the research conducted in the paper conform, in every respect, with the NeurIPS Code of Ethics \url{https://neurips.cc/public/EthicsGuidelines}?
    \item[] Answer: \answerYes{} %
    \item[] Justification: \justificationYes{The research uses publicly available datasets and models under their published terms, plus preference pairs the authors generated from public problems. It involves no human subjects and no personal data.}

\item {\bf Broader impacts}
    \item[] Question: Does the paper discuss both potential positive societal impacts and negative societal impacts of the work performed?
    \item[] Answer: \answerNo{} %
    \item[] Justification: \justificationNo{The paper does not include a broader impacts discussion. This is foundational research on alignment loss dynamics with no direct path to specific negative applications beyond those inherent to LLM alignment generally.}

\item {\bf Safeguards}
    \item[] Question: Does the paper describe safeguards that have been put in place for responsible release of data or models that have a high risk for misuse (e.g., pre-trained language models, image generators, or scraped datasets)?
    \item[] Answer: \answerNA{} %
    \item[] Justification: \justificationNA{The paper does not release new models or datasets; it analyzes existing alignment methods and proposes a loss modification tested on publicly available models and data.}

\item {\bf Licenses for existing assets}
    \item[] Question: Are the creators or original owners of assets (e.g., code, data, models), used in the paper, properly credited and are the license and terms of use explicitly mentioned and properly respected?
    \item[] Answer: \answerYes{} %
    \item[] Justification: \justificationYes{All datasets and models are cited with their original papers and used under their published terms. Datasets: MATH (MIT), GSM8K (MIT), MBPP (CC-BY-4.0), UltraFeedback (MIT), UltraChat-200k (MIT), IFEval (Apache-2.0), AIME 2024 (public competition problems). Models: Qwen3 and Qwen3.5 (Apache-2.0), DeepSeek-R1-0528-Qwen3-8B and DeepSeek-R1-Distill-Qwen-7B (MIT), Ministral-3-8B (Apache-2.0), Mistral-7B (Apache-2.0), OLMo-2-7B (Apache-2.0), Gemma~3 4B (Gemma Terms of Use). The preference pairs and correct/incorrect completions were generated by the authors from these public problems.}

\item {\bf New assets}
    \item[] Question: Are new assets introduced in the paper well documented and is the documentation provided alongside the assets?
    \item[] Answer: \answerNA{} %
    \item[] Justification: \justificationNA{The paper does not release new code, datasets, or model weights.}

\item {\bf Crowdsourcing and research with human subjects}
    \item[] Question: For crowdsourcing experiments and research with human subjects, does the paper include the full text of instructions given to participants and screenshots, if applicable, as well as details about compensation (if any)?
    \item[] Answer: \answerNA{} %
    \item[] Justification: \justificationNA{The paper does not involve crowdsourcing or human subjects research.}

\item {\bf Institutional review board (IRB) approvals or equivalent for research with human subjects}
    \item[] Question: Does the paper describe potential risks incurred by study participants, whether such risks were disclosed to the subjects, and whether Institutional Review Board (IRB) approvals (or an equivalent approval/review based on the requirements of your country or institution) were obtained?
    \item[] Answer: \answerNA{} %
    \item[] Justification: \justificationNA{The paper does not involve human subjects research.}

\item {\bf Declaration of LLM usage}
    \item[] Question: Does the paper describe the usage of LLMs if it is an important, original, or non-standard component of the core methods in this research? Note that if the LLM is used only for writing, editing, or formatting purposes and does \emph{not} impact the core methodology, scientific rigor, or originality of the research, declaration is not required.
    \item[] Answer: \answerNA{} %
    \item[] Justification: \justificationNA{LLMs are the subject of study (as models being aligned) but are not used as a non-standard component of the research methodology itself.}

\end{enumerate}

\end{document}